%% file: main.tex
\documentclass[10pt,journal,twoside]{IEEEtran}
\IEEEoverridecommandlockouts
\usepackage{amsmath,amssymb,amsfonts}
\usepackage{algorithmic}
\usepackage{graphicx}
\usepackage{textcomp}
\usepackage{color}
\usepackage{xcolor}
\usepackage{threeparttable}
\usepackage{tabularx}
\usepackage{booktabs}
\usepackage{caption}
\usepackage{multirow}
\usepackage{colortbl}
\usepackage{subfigure}
\usepackage{pifont}
\usepackage{comment}
\usepackage{stfloats}
\usepackage{bbding}
\usepackage{multirow}
\usepackage{ulem}
\usepackage[colorlinks, linkcolor=magenta, anchorcolor=green, citecolor=blue]{hyperref}
\usepackage[numbers,sort&compress]{natbib}

\usepackage{graphicx,pifont}
\let\oldding\ding
\renewcommand{\ding}[2][1]{\scalebox{#1}{\oldding{#2}}}
\usepackage{setspace}

\usepackage[linesnumbered,ruled]{algorithm2e}
\def\BibTeX{{\rm B\kern-.05em{\sc i\kern-.025em b}\kern-.08em
    T\kern-.1667em\lower.7ex\hbox{E}\kern-.125emX}}

\begin{document}

\onecolumn
\clearpage
\setcounter{page}{1}
\twocolumn

\title{
TinyFormer: Efficient Sparse Transformer Design and Deployment on Tiny Devices
}

\author{Jianlei~Yang,~\IEEEmembership{Senior~Member,~IEEE,}
        Jiacheng Liao,
        Fanding Lei,
        Meichen Liu,
        Lingkun Long, 
        Junyi Chen, \\
        Han Wan,
        Bei Yu,~\IEEEmembership{Senior~Member,~IEEE}
        and~Weisheng~Zhao,~\IEEEmembership{Fellow,~IEEE}

\thanks{Manuscript received on August, 2024, revised on March 2025, accepted on November 2025.
This work is supported in part by the Beijing Natural Science Foundation (Grant No. L243031), the National Key R\&D Program of China (Grant No. 2023YFB4503704 and 2024YFB4505601), and the National Natural Science Foundation of China (Grant No. 62572036). \textit{Corresponding authors are Jianlei Yang and Han Wan}.
}
\thanks{
J. Yang, J. Liao, F. Lei, M. Liu, L. Long, J. Chen and H. Wan are with School of Computer Science and Engineering, Beihang University, Beijing 100191, China, and Qingdao Research Institute, Beihang University, Qingdao 266104, China.
Email: \url{jianlei@buaa.edu.cn}, \url{wanhan@buaa.edu.cn}
}
\thanks{
B. Yu is with Department of Computer Science and Engineering, The Chinese University of Hong Kong, Hong Kong SAR, China.
}
\thanks{
W. Zhao is with Fert Beijing Research Institute, School of Integrated Circuit Science and Engineering, Beihang University, Beijing, 100191, China.
}
}

\maketitle

\input{0-abstract.tex}


\input{1-introduction.tex}
\input{2-related_work.tex}
\input{3-tinyformer.tex}
\input{4-experiment.tex}

\input{5-conclusion.tex}

{
\small
\bibliographystyle{IEEEtran}
\bibliography{ref.bib}
}

\begin{IEEEbiography}[{\includegraphics[width=1in,height=1.25in,clip,keepaspectratio]{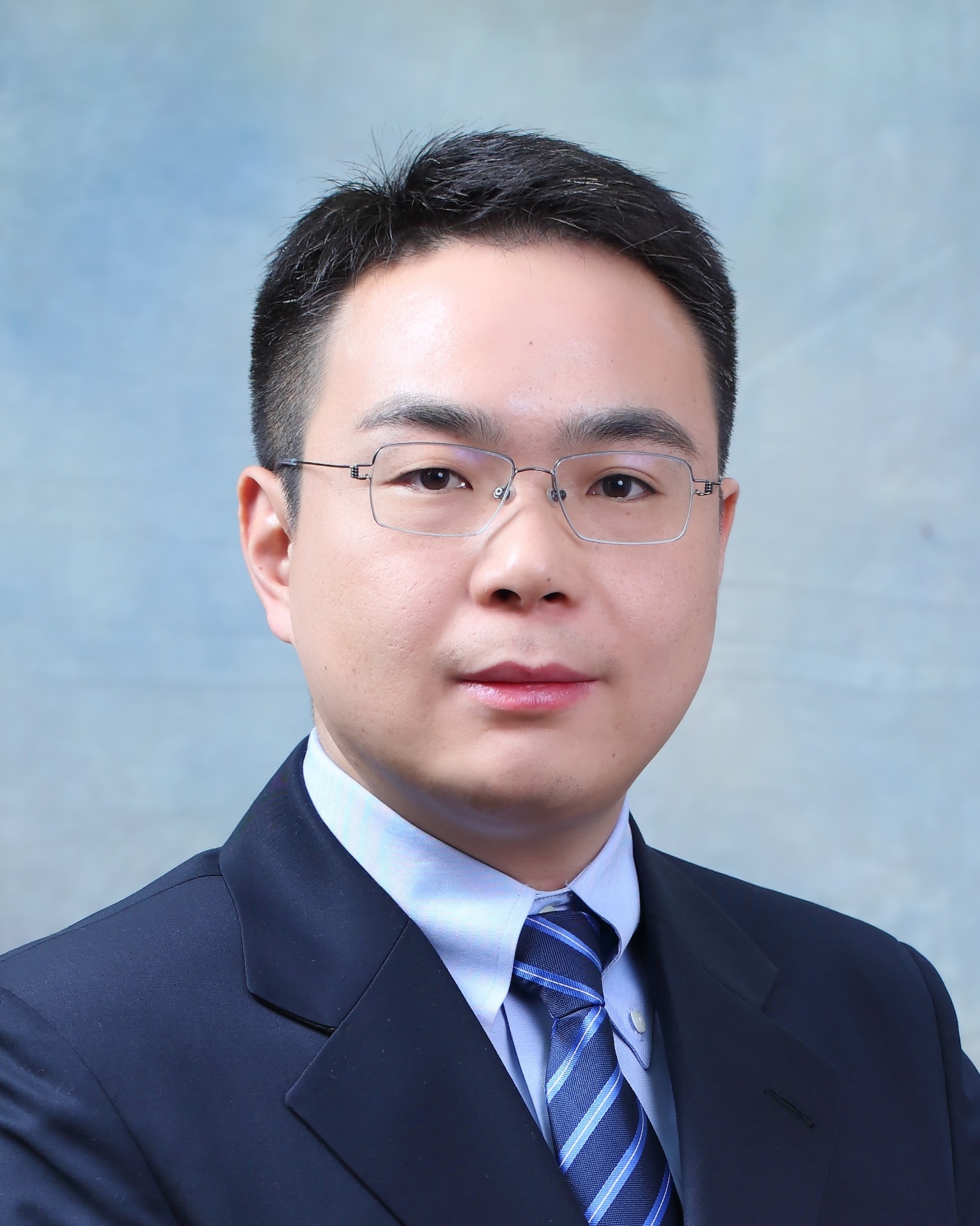}}]{Jianlei Yang}

(S'11-M'14-SM'20) received the B.S. degree in microelectronics from Xidian University, Xi'an, China, in 2009, and the Ph.D. degree in computer science and technology from Tsinghua University, Beijing, China, in 2014.

He is currently a Professor in Beihang University, Beijing, China, with the School of Computer Science and Engineering. From 2014 to 2016, he was a post-doctoral researcher with the Department of ECE, University of Pittsburgh, Pennsylvania, USA.
His current research interests include emerging computer architectures, hardware-software co-design and machine learning systems.

Dr. Yang was the recipient of the First/Second place on ACM TAU Power Grid Simulation Contest in 2011 and 2012. He was a recipient of IEEE ICCD Best Paper Award in 2013, ACM GLSVLSI Best Paper Nomination in 2015, IEEE ICESS Best Paper Award in 2017, ACM SIGKDD Best Student Paper Award in 2020.

\end{IEEEbiography}

\vspace{-8mm}

\begin{IEEEbiography}[{\includegraphics[width=1in,height=1.25in,clip,keepaspectratio]{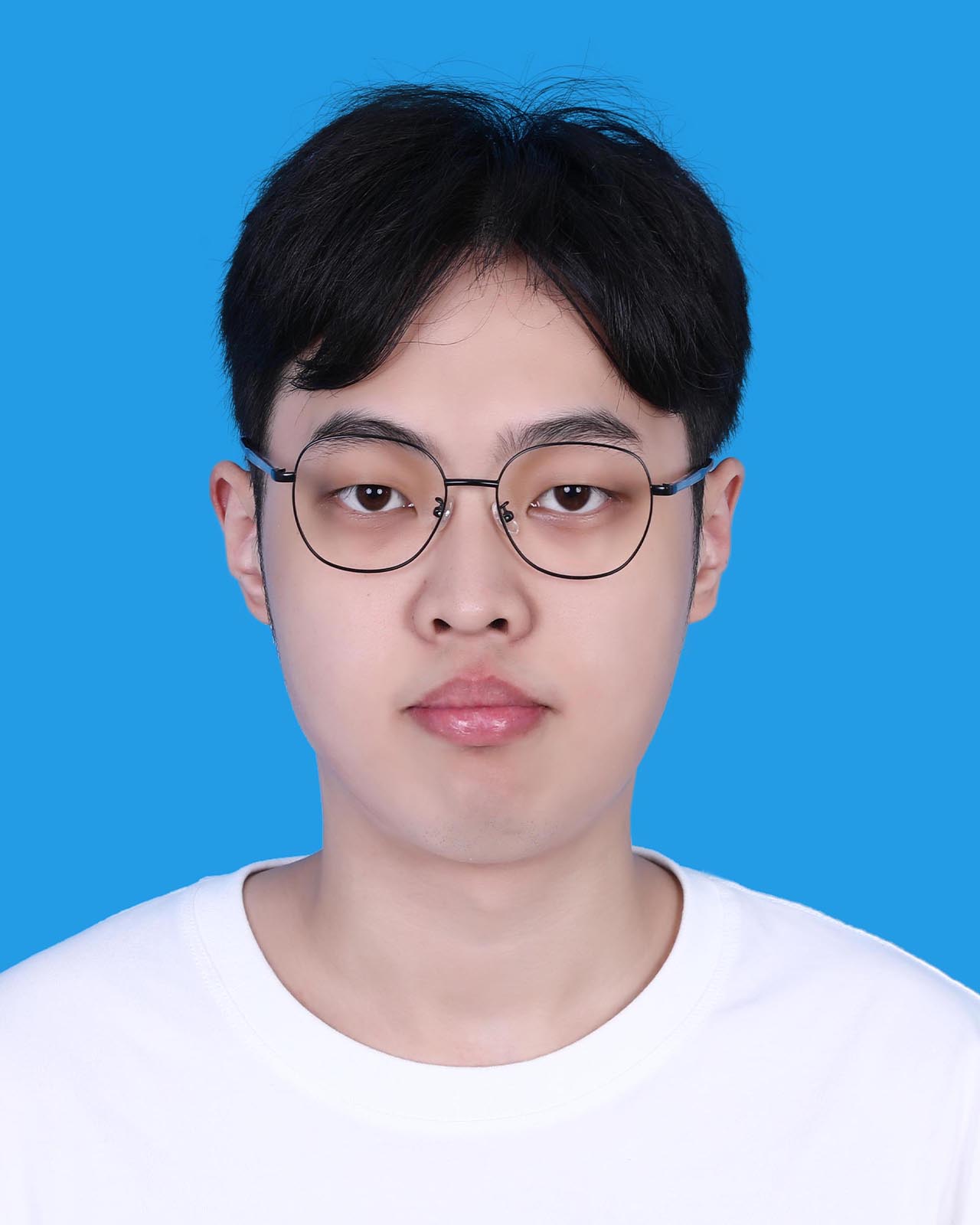}}]{Jiacheng Liao}

received the B.S. degree in Computer Science and Technology from Beihang University, Beijing, China, in 2022, and M.S. degree in Computer Science and Technology in Beihang University, Beijing, China, in 2025. 
His current research interests include TinyML and ML systems.

\end{IEEEbiography}

\vspace{-8mm}

\begin{IEEEbiography}[{\includegraphics[width=1in,height=1.25in,clip,keepaspectratio]{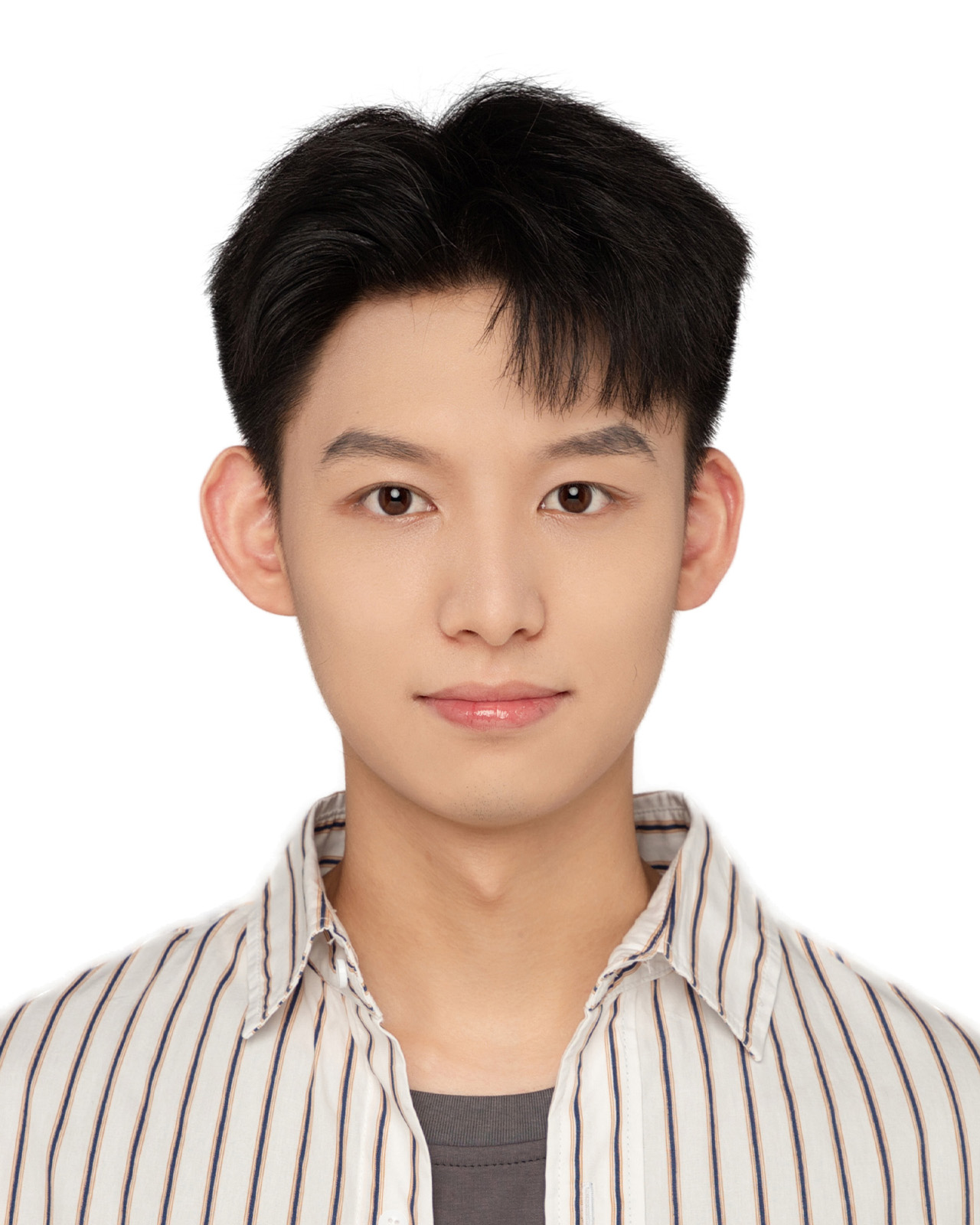}}]{Fanding Lei}

received the B.S. degree in Information and Electronics from Beijing Institute of Technology, Beijing, China, in 2020, and M.S. degree in Computer Science and Technology in Beihang University, Beijing, China, in 2023.
His current research interests include TinyML and ML systems.

\end{IEEEbiography}
\vspace{-8mm}

\begin{IEEEbiography}[{\includegraphics[width=1in,height=1.25in,clip,keepaspectratio]{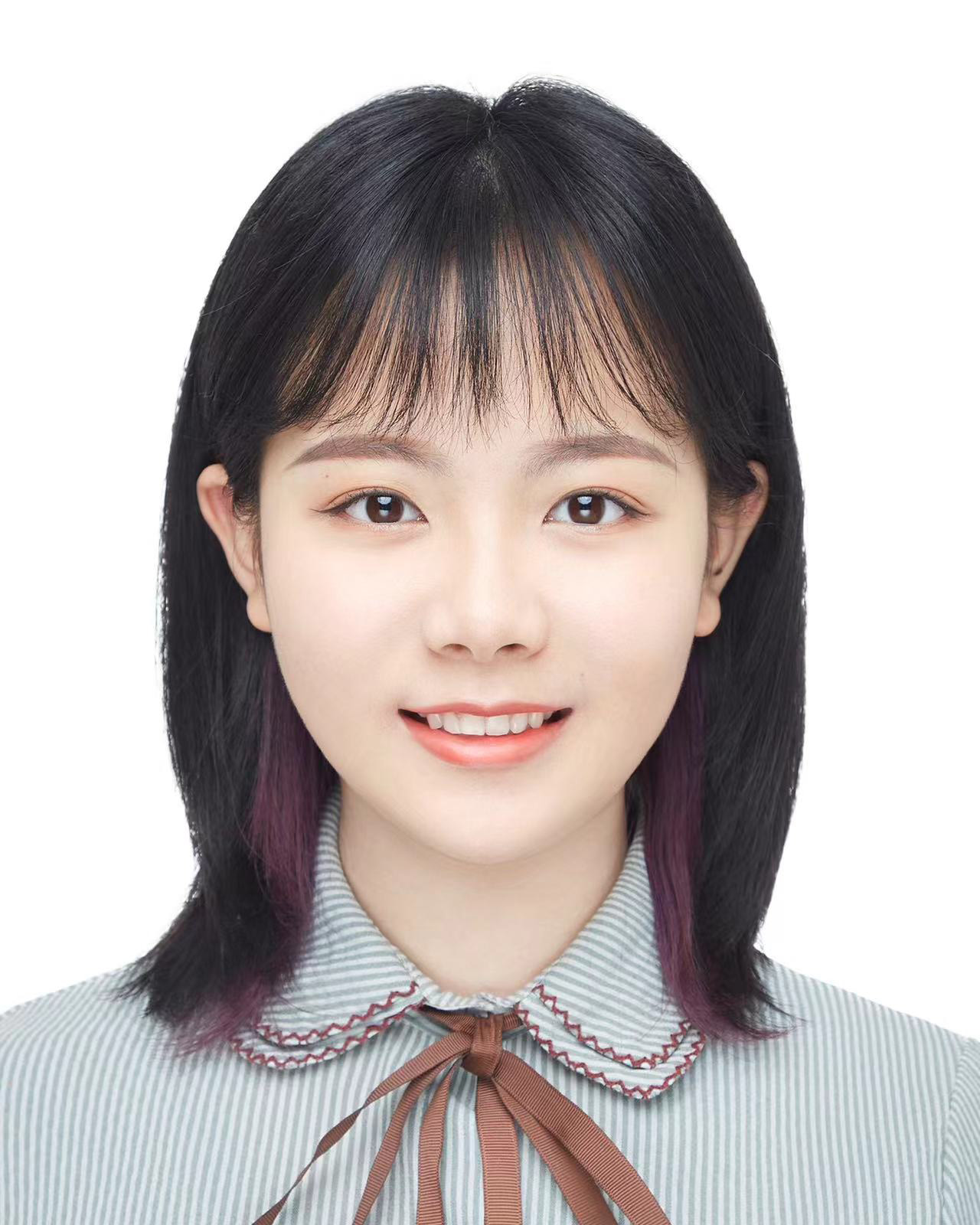}}]{Meichen Liu}

received the B.S. degree in Computer Science and Technology from Beihang University, Beijing, China, in 2021, and M.S. degree in Computer Science and Technology in Beihang University, Beijing, China, in 2024. 
Her current research interests include TinyML and ML systems.

\end{IEEEbiography}

\vspace{-8mm}

\begin{IEEEbiography}[{\includegraphics[width=1in,height=1.25in,clip,keepaspectratio]{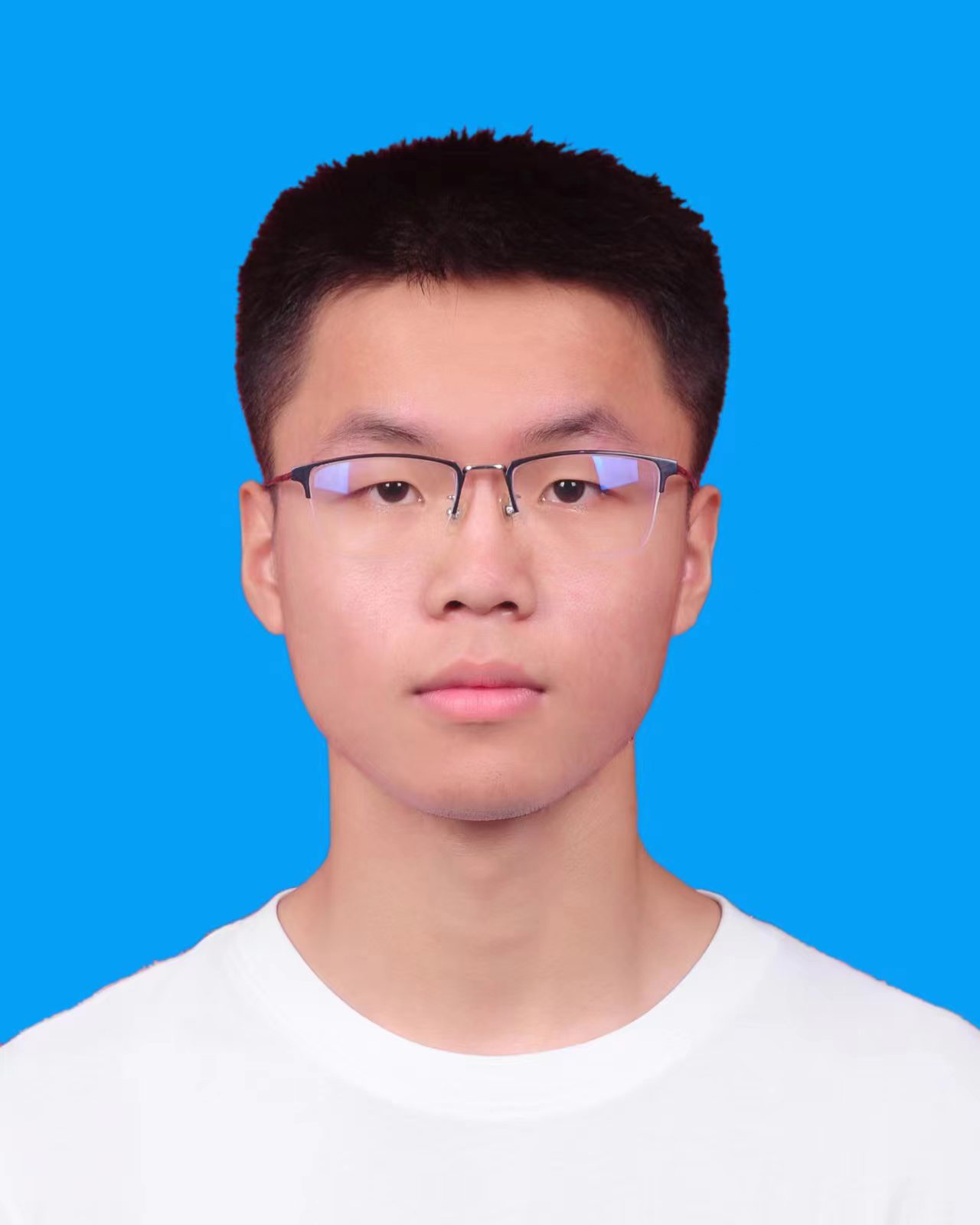}}]{Lingkun Long}

received the B.S. degree in Computer Science and Technology from Beihang University, Beijing, China, in 2024. He is currently working toward the M.S. degree in Computer Science and Technology in Beihang University, Beijing, China.
His current research interests include LLM inference and ML systems.

\end{IEEEbiography}

\vspace{-8mm}

\begin{IEEEbiography}[{\includegraphics[width=1in,height=1.25in,clip,keepaspectratio]{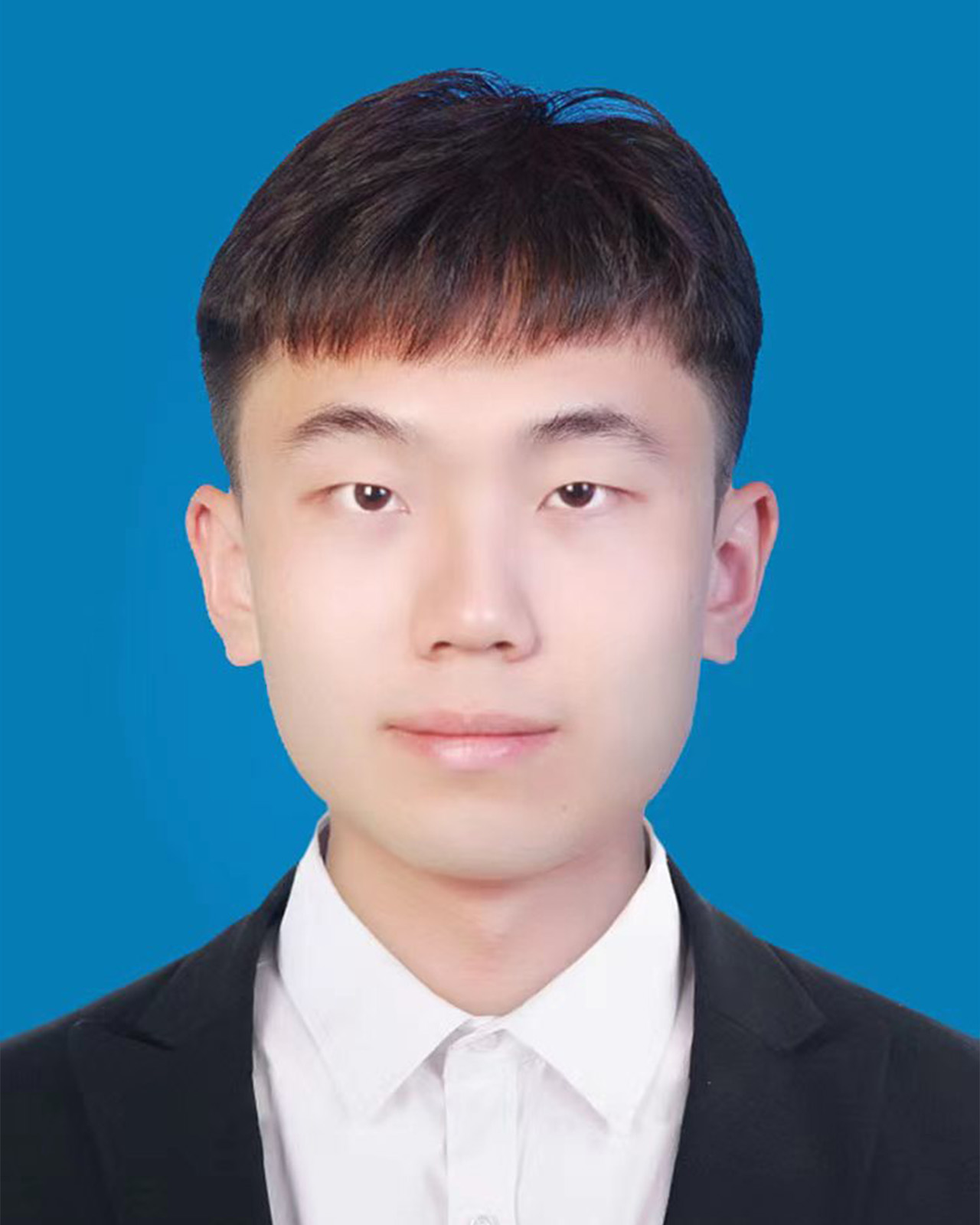}}]{Junyi Chen}

received the B.S. degree in Computer Science and Technology from Beihang University, Beijing, China, in 2024. He is currently working toward the M.S. degree in Computer Science and Technology in Shanghai Jiao Tong University, Shanghai, China. His current research interests include ML systems.

\end{IEEEbiography}

\vspace{-8mm}

\begin{IEEEbiography}[{\includegraphics[width=1in,height=1.25in,clip,keepaspectratio]{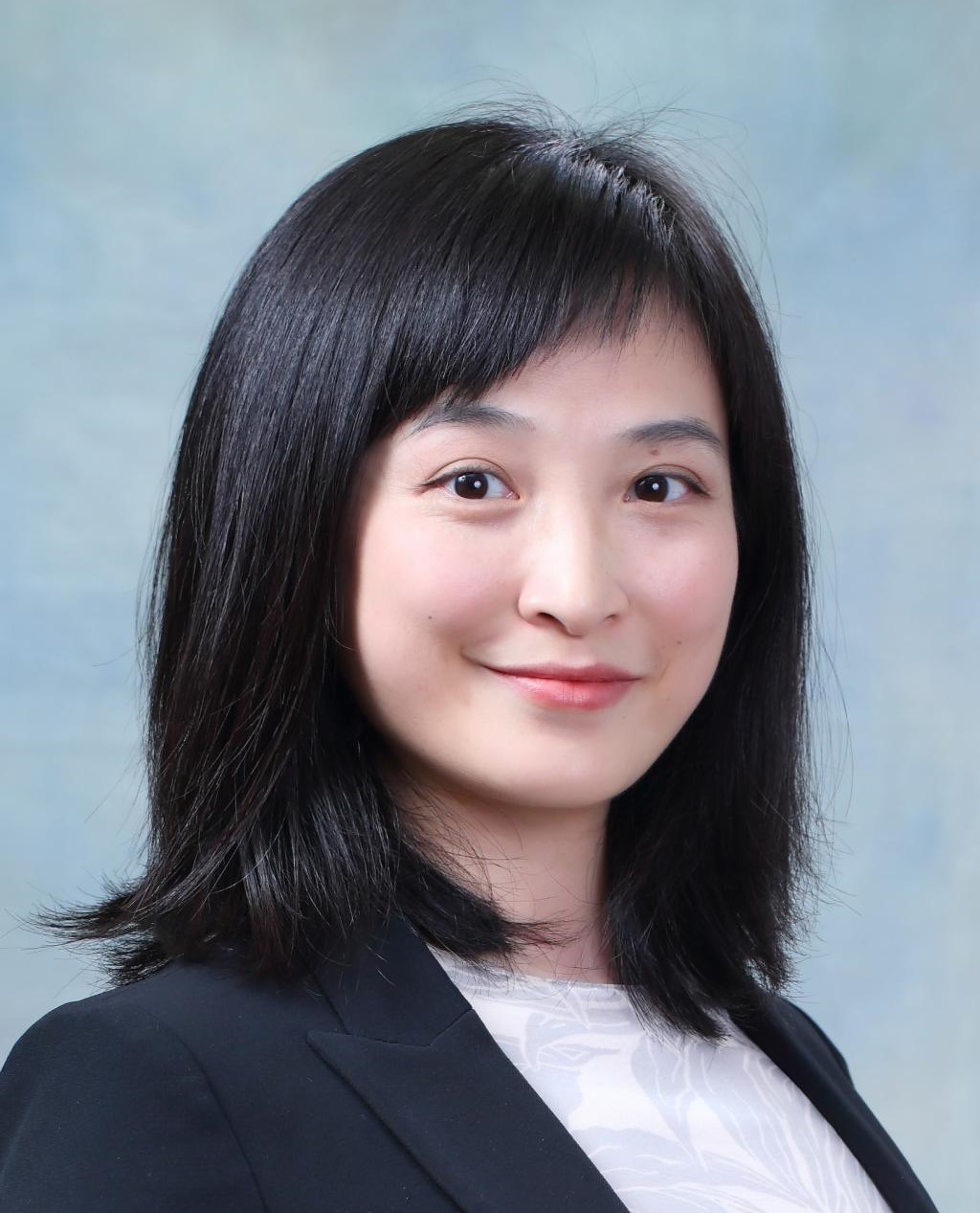}}]{Han Wan}

received the B.S. degree and Ph.D. degree in Computer Science and Technology from Beihang University, Beijing, China, in 2003 and 2011, respectively.
She is currently an Associate Professor with the School of Computer Science and Engineering at Beihang University.
From 2015 to 2016, she was a visiting scholar with the Education Research Group, Massachusetts Institute of Technology (MIT).

Her research interests include computer architectures and systems, educational data mining.

\end{IEEEbiography}

\vspace{-8mm}

\begin{IEEEbiography}[{\includegraphics[width=1in,height=1.25in,clip,keepaspectratio]{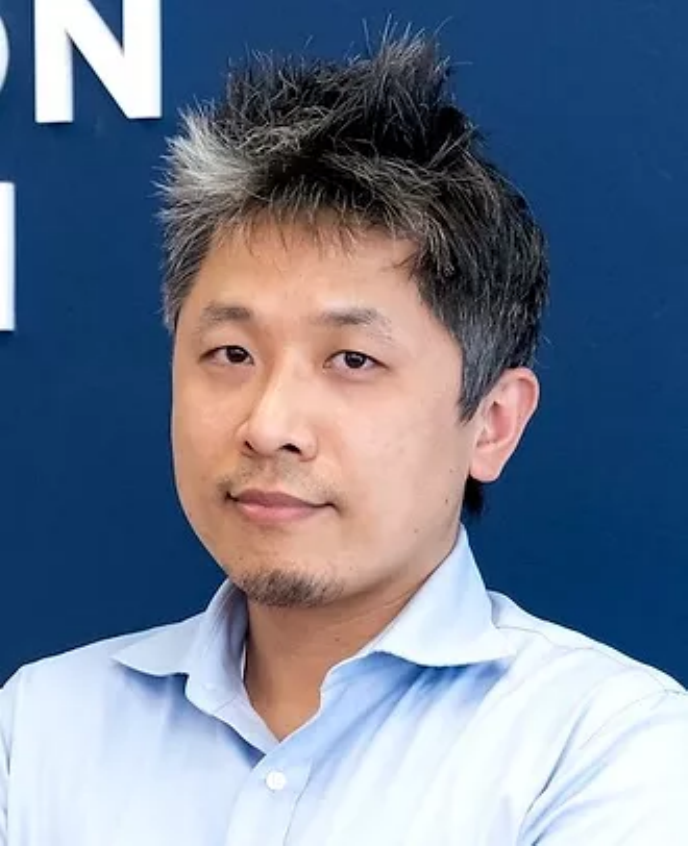}}]{Bei Yu} 

(M'15-SM'22) received the Ph.D. degree from The University of Texas at Austin in 2014. He is currently a Professor in the Department of Computer Science and Engineering, The Chinese University of Hong Kong.

He has served as TPC Chair of ACM/IEEE Workshop on Machine Learning for CAD, and in many journal editorial boards and conference committees. He is Editor of IEEE TCCPS Newsletter. He received nine Best Paper Awards from DATE 2022, ICCAD 2021 \& 2013, ASPDAC 2021 \& 2012, ICTAI 2019, Integration, the VLSI Journal in 2018, ISPD 2017, SPIE Advanced Lithography Conference 2016, and six ICCAD/ISPD contest awards.

\end{IEEEbiography}

\vspace{-8mm}

\begin{IEEEbiography}[{\includegraphics[width=1in,height=1.25in,clip,keepaspectratio]{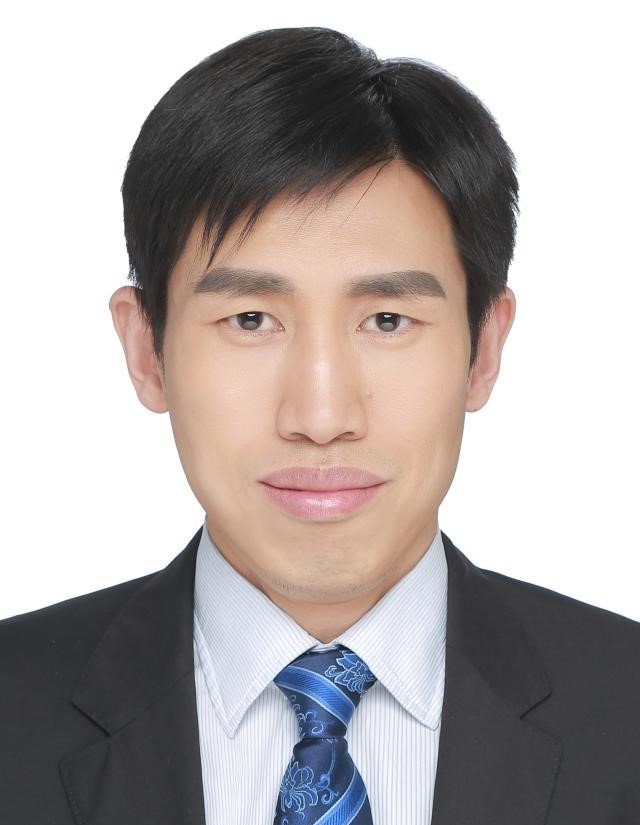}}]{Weisheng Zhao}

(Fellow, IEEE) received the Ph.D. degree in physics from the University of Paris Sud, Paris, France, in 2007.

He is currently a Professor with the School of Integrated Circuit Science and Engineering, Beihang University, Beijing, China. In 2009, he joined the French National Research Center, Paris, as a Tenured Research Scientist. Since 2014, he has been a Distinguished Professor with Beihang University. He has published more than 200 scientific articles in leading journals and conferences, such as \textit{Nature
Electronics}, \textit{Nature Communications}, \textit{Advanced Materials}, IEEE Transactions, ISCA, and DAC. His current research interests include the hybrid integration of nanodevices with CMOS circuit and new nonvolatile memory (40-nm technology node and below) like MRAM circuit and architecture design.

Prof. Zhao was the Editor-in-Chief for the {\sc{IEEE Transactions on Circuits and System I: Regular Paper}} from 2020 to 2023.

\end{IEEEbiography}

\end{document}

%% file: 0-abstract.tex
\begin{abstract}

Developing deep learning models on tiny devices (e.g. Microcontroller units, MCUs) has attracted much attention in various embedded IoT applications.
However, it is challenging to efficiently design and deploy recent advanced models (e.g. transformers) on tiny devices due to their severe hardware resource constraints.
In this work, we propose \textit{TinyFormer}, a framework specifically designed to develop and deploy resource-efficient transformer models on MCUs.
TinyFormer consists of \textit{SuperNAS}, \textit{SparseNAS}, and \textit{SparseEngine}. 
Separately, SuperNAS aims to search for an appropriate supernet from a vast search space.
SparseNAS evaluates the best sparse single-path transformer model from the identified supernet. 
Finally, SparseEngine efficiently deploys the searched sparse models onto MCUs.
To the best of our knowledge, SparseEngine is the first deployment framework capable of performing inference of sparse transformer models on MCUs.
Evaluation results on the CIFAR-10 dataset demonstrate that TinyFormer can design efficient transformers with an accuracy of $96.1\%$ while adhering to hardware constraints of $1$MB storage and $320$KB memory.
Additionally, TinyFormer achieves significant speedups in sparse inference, up to $12.2\times$ comparing to the CMSIS-NN library.
TinyFormer is believed to bring powerful transformers into TinyML scenarios and to greatly expand the scope of deep learning applications.

\end{abstract}


%% file: 1-introduction.tex
\section{Introduction}

\begin{figure}[t]
    \centering
    \includegraphics[width=\columnwidth]{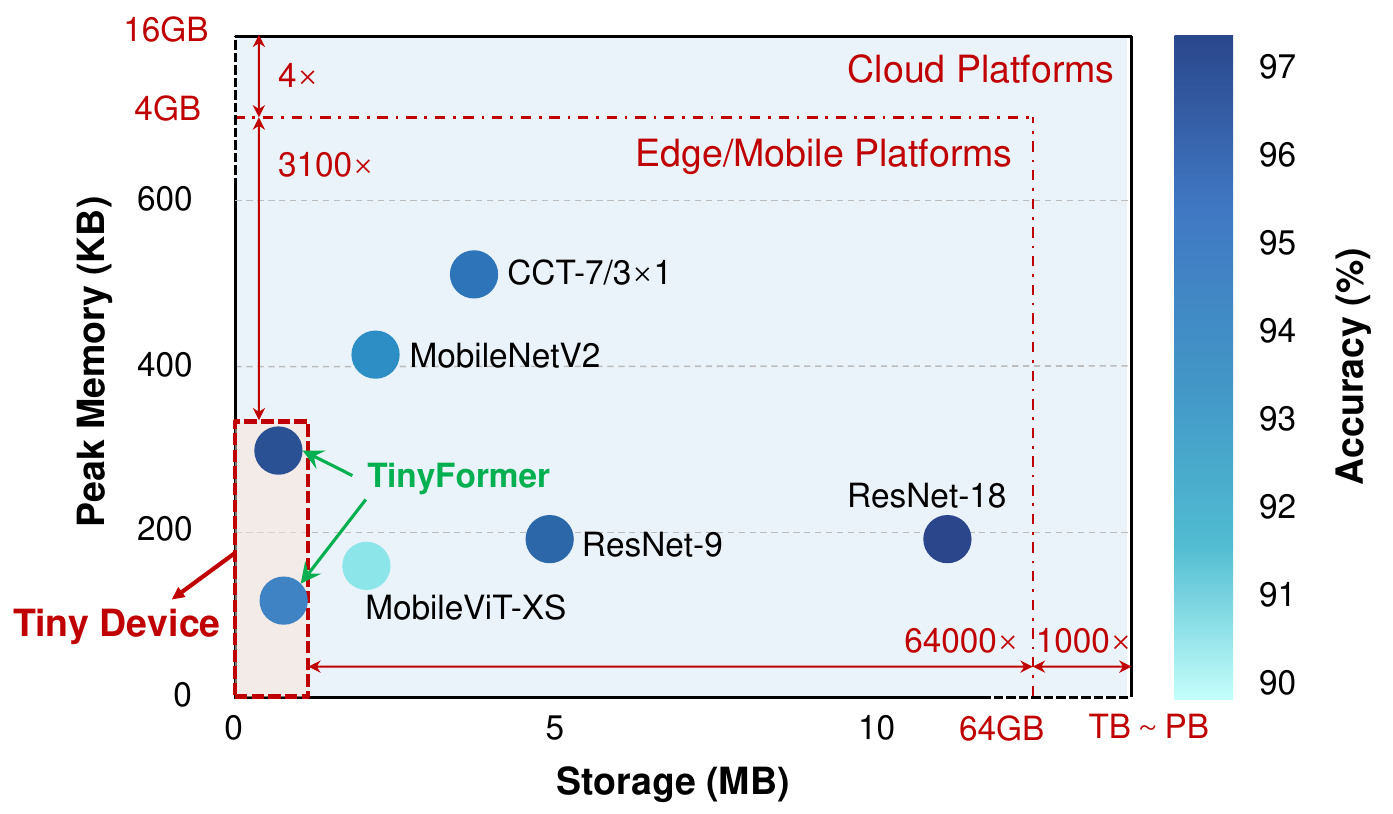}
    \caption{
        Accuracy and resource usage comparison among different compact models when evaluated on CIFAR-10.
        Tiny devices only have MB-level storage and KB-level memory, which is a huge difference between the resources available on edge or cloud platforms.
    }
    \label{fig:fig-intro-overview}
\end{figure}

\IEEEPARstart{A}{s} IoT applications are becoming increasingly popular recently, microcontroller units (MCUs) have received extensive attention among various kinds of application scenarios.
These low-cost, low-power tiny devices are wildly used in plug-and-play scenarios with extreme resource constraints.
The devices are usually deployed near the sensor end, gathering the freshest data once produced.
Accordingly, Tiny Machine Learning (TinyML) is a growing field in computer science, aiming to apply machine learning technology on MCUs, thereby enabling various applications \cite{shafique2021tinyml}.
Several well-established TinyML applications, such as \textit{Keyword Spotting} \cite{zhang2017hello}, \textit{Anomaly Detection} and \textit{Raise to Wake}, only involve simple machine learning algorithms.
Some higher-end applications, such as \textit{Wildlife Detection} and \textit{Food Edibility Detection}, usually require powerful deep learning models \cite{kallimani2023tinyml}.
However, most of these scenarios only have MCUs available to be exploited, which poses new challenges to TinyML. 

Bringing powerful deep neural networks to MCUs can greatly expand the scope of deep learning applications \cite{banbury2020benchmarking, banbury2021micronets}.
However, the available resources of MCUs are strictly limited.
For example, ARM Cortex-M7 has only $1$MB storage (Flash) and $320$KB memory (SRAM)
The resources of ARM Cortex-M7 are even less than that of mobile devices (such as mobile phones, Raspberry Pi) which are up to GB-level.
As shown in Fig. \ref{fig:fig-intro-overview}, there is a large gap between the required resources of deep learning models and the available hardware capacities of tiny devices.
To deploy ResNet-18 \cite{He2016CVPR} (with $11$M parameters) on MCUs, at least $90\%$ of weights have to be shrunk, which leads to significant accuracy degradation.
Moreover, weight pruning does not reduce the peak memory of deep learning models.
It is necessary to redesign the neural network to reduce the peak memory.
Therefore, it is difficult to deploy powerful models on such resource-constrained devices.

Recently, transformers have achieved great performance in various fields, including computer vision, speech recognition, and natural language processing \cite{VaswaniSPUJGKP17, GulatiQCPZYHWZW20, Dosovitskiy2021AnII}. 
Deploying these powerful transformer models on MCUs can be exciting for satisfying the requirement of high-demanding scenarios in TinyML. 
However, transformers contain a large number of parameters.
Even the lightweight transformer models \cite{mehta2021mobilevit, zhang2022edgeformer} can not satisfy the strict resource constraints. 
Deploying powerful transformer models on edge devices or even MCUs remains difficult.

Aiming to bring powerful transformers to MCUs, TinyFormer is proposed as an efficient framework to design and deploy sparse transformers on resource-constrained devices.
To be noticed, a sparse transformer in this paper refers to a hybrid model that consists of transformer encoders and convolution layers, with a weight pruning strategy.
TinyFormer consists of SuperNAS, SparseNAS, and SparseEngine.
SuperNAS aims to produce an appropriate supernet from a large search space for further searching.
SparseNAS performs single-path model searching in the supernet and model compressing for evaluating hardware constraints and accuracy.
SparseEngine automatically deploys and optimizes the compressed models with the highest accuracy on targeted MCUs.
The main contributions of this paper are as follows:
\begin{itemize}
    \item TinyFormer is proposed as an efficient framework to develop transformers on resource-constrained devices.
    TinyFormer brings powerful transformers into TinyML scenarios by making it extremely small and efficient on MCUs.
    \item With the joint search and optimization of sparsity configuration and model architecture, TinyFormer produces a sparse transformer with the best accuracy while satisfying hardware constraints.
    \item We propose SparseEngine, an automated deployment tool for sparse transformers-based hybrid models.
    To the best of our knowledge, it is the first deployment tool capable of performing sparse inference for transformers, with a guaranteed latency on targeted MCUs.
\end{itemize}

\textcolor{black}{With the cooperation of SuperNAS, SparseNAS, and SparseEngine, TinyFormer brings powerful transformers into resource-constrained devices, and enables a faster sparse inference process compared with existing inference engines.} 
Experimental results on CIFAR-10 show that TinyFormer could achieve an accuracy of $96.1\%$, with an inference latency of $3.9$ seconds running on STM32F746. 
Compared to the light-weight transformer CCT-7/3$\times$1 \cite{hassani2021escaping}, TinyFormer achieves higher accuracy improvement while saving $74\%$ storage.
Benefiting from the automated SparseEngine, TinyFormer could obtain up to $12.2\times$ speedup in sparse model inference compared to CMSIS-NN \cite{lai2018cmsis}.

The rest of the paper is organized as follows. 
Section \ref{sec:related_work} reviews the related background and provides our motivations.
Section \ref{sec:tinyformer} demonstrates the details of the TinyFormer framework. 
Experimental results are presented in Section \ref{sec:experiment} and conclusions are given in Section \ref{sec:conclusion}.

%% file: 2-related_work.tex
\section{Related Works and Motivations} \label{sec:related_work}


Aiming to bring transformers to TinyML scenarios, the following issues have to be comprehensively considered. 
Firstly, the transformer architecture should be light-weighted to satisfy the extremely demanding resource constraints. 
Moreover, the sparsity configuration and architecture of the model have coupled effects on accuracy. 
Thus, we are supposed to consider these coupled effects in the model architecture exploration and compression stage. 
Finally, it is essential to provide specialized deployment support for targeting to MCUs. 
These representative investigations motivate us to enable powerful deep-learning models on MCUs through various technological paradigms. 
Our three primary motivations are listed below:

\vspace{1mm}
\textbf{Motivation \raisebox{-0.1em}{\ding[1.3]{182}}}: Design light-weight transformer for MCU.



Existing deep learning models on MCU for computer vision tasks are mostly CNNs.
Aiming to meet the resource constraints, researchers use mixed precision quantization to deploy CNN models on MCUs \cite{Rusci2020MemoryDrivenML, Rusci2020LeveragingAM}.
MCUNet \cite{lin2020mcunet, lin2021mcunetv2} is a system-algorithm co-design framework to search and deploy extremely tiny models on the MCU platform. MCUNet aims to find the models with low resources required, and their  architectures are derived from the basic structures of CNNs. 
Some researchers even attempt to train deep learning models on edge devices  \cite{lin2022ondevice}.


Besides CNNs, Transformers have demonstrated success in a large amount of downstream tasks \cite{VaswaniSPUJGKP17, GulatiQCPZYHWZW20, Dosovitskiy2021AnII} including computer vision ones.
However, there are fewer works to deploy Transformer on MCUs due to the large amount of parameters and high peak memory during inference.
Even the light-weight transformers require resources far beyond the upper limit of MCU constraints \cite{mehta2021mobilevit,hassani2021escaping,zhang2022edgeformer}.
Pulp-transformer \cite{Burrello21pulp} introduces a library of attention kernels to accelerate transformers' inference.
\textcolor{black}{MCUFormer \cite{Liang2023mcuformer} brings transformer on MCU and achieves high accuracy in image classification on ImageNet-1k.
However, MCUFormer is mainly focused on optimizing the computational graph to accelerate the inference process with dense models.
Applying pruning methods and exploiting the sparsity of the model can further improve the performance of transformers on MCU.}

It is still challenging to deploy very small transformers on MCUs.
From our perspectives, these efforts will bring transformers into TinyML scenarios, enhancing the applicability of numerous tiny devices.

\vspace{1mm}
\textbf{Motivation \raisebox{-0.1em}{\ding[1.3]{183}}}: Joint search with model architecture and sparsity configuration.


Neural Architecture Search (NAS) is an advanced approach in automatically model architecture design \cite{ZophL17} and achieves remarkable results in various fields. 
Since model deployment is bounded by hardware constraints (including storage, memory and inference latency), Hardware-Aware Neural Architecture Search (HW-NAS) algorithm is proposed to adopt NAS algorithms for target hardware devices.
Most HW-NAS approaches are targeting GPU \cite{zhou2023hardware}, mobile devices \cite{WuDZWSWTVJK19, CaiZH19} or custom hardware \cite{JiangYSZGDSH20}.

Most NAS approaches aim to search for a dense model for better performance.
However, the lack of considering model sparsity in NAS limits the potential benefits of efficient TinyML exploration. 
SpArSe proposed a sparse architecture search algorithm for resource-constrained MCUs \cite{fedorov2019sparse}. 
However, SpArSe optimizes the network's morphology and performs searching and pruning on respective stages. 
Hence, it ignores the coupled influence of pruning parameters and network architecture on model accuracy. 
Aiming to fully consider the interaction between sparsity configuration and dense single-path model, we add pruning steps in two-stage NAS for validation.

\vspace{1mm}
\textbf{Motivation \raisebox{-0.1em}{\ding[1.3]{184}}}: Deploy efficient sparse models on MCU.

Model compression can significantly reduce the model size while maintaining the accuracy. 
Accordingly, deploying compressed models on MCUs requires particular support from deployment tools. 
For example, deployment tools should support sparse model storing and execution to accommodate their extremely severe resource constraints.
Existing deployment tools and libraries for MCUs include TensorFlow Lite Micro \cite{Abadi2016TensorFlowAS}, CMSIS-NN \cite{lai2018cmsis}, MicroTVM \cite{Chen2018TVMAA}, CMix-NN \cite{capotondi2020cmix}.
TinyEngine \cite{lin2020mcunet} is a code generator-based library to save the resource consumed by the interpreter during inference.

However, these frameworks are mainly targeted at small-scale computing by utilizing the locality in dense form for acceleration. 
Support of sparse coding and inference enables deployment tools to accommodate larger deep-learning models.
To the best of our knowledge, there are few deployment tools or libraries that support sparse model inference, limiting the scale and performance of models.
To further exploit the advantage of sparsity, a specialized deployment tool should be developed for sparse model inference on MCUs.

%% file: 3-tinyformer.tex
\section{TinyFormer: A Framework of Resource-Efficient Model Searching and Deployment}
\label{sec:tinyformer} 

\begin{figure}[t]
    \centering
    \includegraphics[width=1\columnwidth]{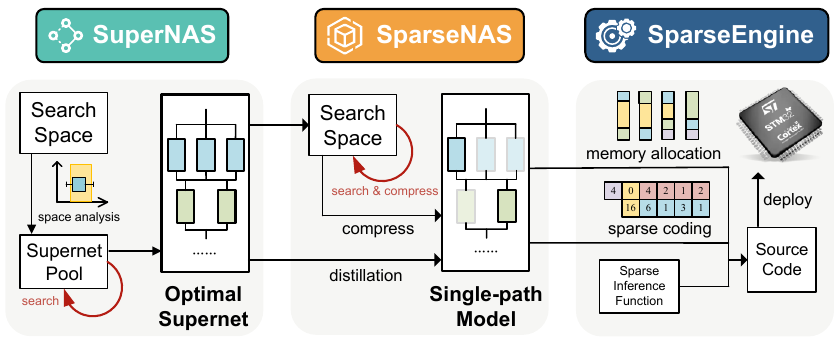}
    \caption{TinyFormer is a hardware-aware framework. SuperNAS is co-designed with SparseNAS to produce sparse models with transformers under resource limits. SparseEngine enables sparse inference on MCUs.}
    \label{fig:fig-tinyformer-framework}
\end{figure}

Based on these motivations, we propose TinyFormer, a resource-efficient framework to design and deploy sparse transformer-based hybrid models on resource-constrained devices.
\textcolor{black}{
In Sec. \ref{sec:overview}, we provide a macro view of the TinyFormer's architecture. 
In the rest of the subsection, we discuss about the key components of TinyFormer: SuperNAS, SparseNAS, and SparseEngine.
We present essential algorithms in SuperNAS and SparseNAS and illustrate how they corporate to search the best sparse model in Sec. \ref{sec:supernas} and Sec. \ref{sec:sparsenas}.
The implementation method of SparseEngine and the procedure of sparse inference is provided in Sec. \ref{sec:sparseengine}.}

\subsection{Overview} \label{sec:overview}

As shown in Fig.~\ref{fig:fig-tinyformer-framework}, TinyFormer consists of three parts: SuperNAS, SparseNAS, and SparseEngine.
SuperNAS aims to automatically find an appropriate supernet in a large search space $\mathbb{S}_{sup(er)}$. 
In this work, a supernet is built as a pre-trained over-parameterized model, where the following single-path models are sampled from the supernet. 
SparseNAS is adopted to find a sparse transformer from the supernet.
SparseNAS searches for a single-path model in the supernet with a set of sparse configurations in search space $\mathbb{S}_{spa(rse)}$.
Weight pruning is performed in \texttt{Conv2d} (convolution in 2d) and \texttt{Linear} operators, and full-integer quantization with \texttt{INT8} format is applied on all layers. 
Finally, SparseEngine automatically generates binary code on STM32 MCUs with several functional implementations.
SparseEngine deploys the obtained sparse model to MCUs, enabling sparse inference to save hardware resources.

\begin{figure}[t]
    \centering
    \includegraphics[width=1.0\columnwidth]{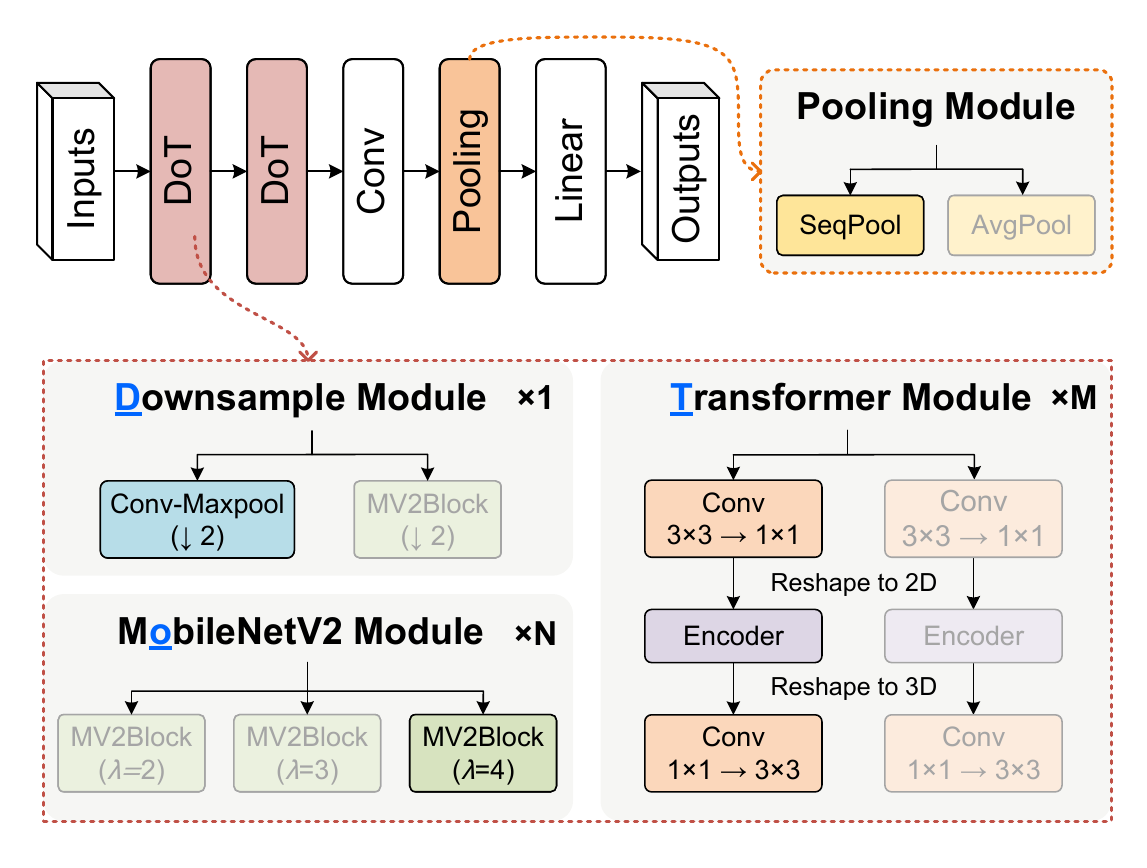}
    \caption{Our supernet architecture design.
    We design four types of choice modules: \textit{Downsample Module}, \textit{MobileNetV2 Module}, 
    \textit{Transformer Module} and
    \textit{Pooling Module}.
    Each choice module contains $2$ or $3$ architecture candidates inside.
    The single-path model is sampled from the supernet with only one architecture candidate invoked per choice module.
    }
    \label{fig:fig-tinyformer-supernet}
\end{figure}

\subsection{SuperNAS: Supernet Architecture Search} \label{sec:supernas}

When designing a supernet structure to search models and deploy on MCUs, we are facing a trade-off between model sparsity and capacity. 
The accuracy of the smaller dense models is limited by capacity, while pruning with higher sparsity on the larger models could lead to a drastic drop in accuracy. 
The balance between model sparsity and capacity should shift according to hardware resource constraints, challenging the design of the supernet.

Therefore, we propose SuperNAS to automatically design the supernet.
SuperNAS searches appropriate supernet in search space $\mathbb{S}_{sup}$. 
Appropriate supernet should satisfy two following conditions. 
Firstly, most of the sparse models obtained from supernet are supposed to satisfy the resource constraints strictly. 
Secondly, the average validation accuracy of these models should be achieved as high as possible.

With the design of the search space for supernet ($\mathbb{S}_{sup}$), SuperNAS first analyses the probability of accepting the designed search space.
Then SuperNAS randomly samples supernet configurations from search space and evaluates the average accuracy of the sparse single-path models in the supernet.
The supernet with the best accuracy will be sent to SparseNAS for further search.

\begin{table}[t]
\renewcommand\arraystretch{1.25}
\caption{Search space design in Sec. \ref{sec:supernas} and \ref{sec:sparsenas}. Configurations of supernet are sampled in $\mathbb{S}_{sup(er)}$, and configurations of the single-path model is sampled in $\mathbb{S}_{spa(rse)}$ with a given supernet. MHSA is the abbreviation of Multi-head Self Attention.}
\label{tab:config}
\begin{center}
\scalebox{0.85}{
\begin{tabular}{l|ll}
\hline
Symbol & Invovled Configurations & Choices\\
\hline
\multirow{6}{*}{$\mathbb{S}_{sup(er)}$} & Number of heads in MHSA\tnote{1} & [1, 2, 4]\\
 & Number of MobileNetV2s in \texttt{DoT}  & [1, 2]\\
 & Number of Transformers in \texttt{DoT} & [1, 2]\\
  & Embedding dimension size in \texttt{DoT1} & [32, 44]\\
  & Embedding dimension size in \texttt{DoT2} & [96, 128, 256]\\ 
  & Dimension of the Last Linear & [512, 768, 1024]\\
\hline
\multirow{3}{*}{$\mathbb{S}_{spa(rse)}$} & Single-path choice of module & Shown in Fig. \ref{fig:fig-tinyformer-supernet}\\
 & Pruning block size & [2, 4]\\
 & Pruning sparsity & [0, 0.4$\sim$0.8:0.1]\\
\hline
\end{tabular}}
\end{center}
\end{table}

\subsubsection{Supernet Architecture design}

The search space of SuperNAS $\mathbb{S}_{sup}$ contains hyper-parameters that are related to the supernet architecture design, shown in Tab. \ref{tab:config}. 
The supernet architecture design is shown in Fig. \ref{fig:fig-tinyformer-supernet}, using the same approaches as the single-path-one-shot (SPOS) \cite{guo2020single}.
Supernet contains choice modules in different branches, and the single-path model is sampled by selecting a choice module at each branch.
We build supernet architecture based on four types of choice modules: \textit{Downsample module}, \textit{MobileNetV2 module}, \textit{Transformer module} and \textit{Pooling module}. 
The four types of modules contain multiple different architecture candidates respectively. 
\textcolor{black}{The condidate parameter set in $\mathbb{S}_{sup}$ is determined based on emperical enginerring practices, and takes multiple aspects into account, including memory usage, storage usage, hardware efficiency, and the trade-off between searched results and the search time in total.}

In Fig. \ref{fig:fig-tinyformer-supernet}, \texttt{Conv} represents a standard $3$$\times$$3$ convolution with $1$$\times$$1$ padding unless stated otherwise.
In \textit{Downsample module}, \texttt{Conv-Maxpool} refers to a standard convolution following a $2\times2$ max pooling.
Architecture candidates that perform downsampling are tagged with $\downarrow 2$.
\textit{MobileNetV2 module} refers to inverted residual block in \cite{Sandler2018MobileNetV2IR} with expansion factor $\lambda$.
In \textit{Transformer module}, \texttt{Conv} $3$$\times$$3$$\rightarrow$$1$$\times$$1$ is a standard $3$$\times$$3$ convolution following a $1$$\times$$1$ convolution with no padding.
\texttt{Conv} $1$$\times$$1$$\rightarrow$$3$$\times$$3$ is similar to the expression above.
Encoder is a standard transformer encoder similar to \cite{Dosovitskiy2021AnII}, with ReLU \cite{agarap2018deep} as the activation operation for efficient calculation on MCUs.
\texttt{SeqPool} and \texttt{AvgPool} in \textit{Pooling module} indicate sequence pooling in \cite{hassani2021escaping} and $2$$\times$$2$ average pooling, respectively.



Unlike CNNs, transformers lack spatial inductive biases and rely heavily on massive datasets for large-scale training.
Therefore, we insert \textit{MobileNetV2 module} before \textit{transformer module} to address this issue.
Referring to MobileViT \cite{mehta2021mobilevit}, we insert \texttt{Conv} $3$$\times$$3$$\rightarrow$$1$$\times$$1$ and \texttt{Conv} $1$$\times$$1$$\rightarrow$$3$$\times$$3$ before- and after- the encoder instead of positional encoding.

We take \texttt{DoT}, an architecture stacked by \textit{\underline{D}ownsample module}, \textit{M\underline{o}bileNetV2 module} and \textit{\underline{T}ransformer module}, as the basic layer of the supernet.
\textit{MobileNetV2 module} and \textit{Transformer module} in \texttt{DoT} structure are repeatable, which makes \texttt{DoT} a more flexible feature extraction architecture.
The supernet architecture adopts two \texttt{DoT}s as the backbone, followed by some post-processing operators.  

\begin{algorithm}[t]
\setstretch{1.1} 
\textbf{Input:} Search space $\mathbb{S}_{sup}$,$\mathbb{S}_{spa}$, memory limit ${\cal L}_{m}$ and storage limit ${\cal L}_{s}$, lower bound ratio $\lambda_{lo}$ and upper bound ratio $\lambda_{up}$, iteration count ${\cal T}_{sup}$. \\
\textbf{Output:} Probability to accept $\mathbb{S}_{sup}$ for search space. \\
$[n_{accpt}, n_{eval}] \leftarrow [0, 0]$\\
\For{$i=1$ \bf{to} ${\cal T}_{sup}$}{
    $c_{sup}, c_{spa}$ $\leftarrow$ $RandomSample(\mathbb{S}_{sup}, \mathbb{S}_{spa})$\\

    ${\cal M}^i$ $\leftarrow$  $CreateModel(c_{sup}, c_{spa})$\\
    
    $[l_m^i,l_s^i]$ $\leftarrow$ $ResourceEval({\cal M}^i)$ \\
    
    \If{$l_m^i$ $\leq$ ${\cal L}_{m}$ \bf{and} $l_s^i$ $\leq$ ${\cal L}_{s}$}{
        $n_{eval}$ $\leftarrow$ $n_{eval}+1$\\
        $N_{param}^i$ $\leftarrow$ $ParamsEval({\cal M}^i)$\\
        \If{$\lambda_{lo}{\cal L}_{m}$ $\leq$ $N_{param}^i$ $\leq$ $\lambda_{up}{\cal L}_{m}$}
        {$n_{accpt}$ $\leftarrow$ $n_{accpt}+1$\\}
    }
    
}
\textbf{return} $n_{accpt}/n_{eval}$
\caption{Search Space Analysis in SuperNAS}
\label{alg:search_space}
\end{algorithm}

\subsubsection{Search Space Analysis}
\label{subsubsec:searchspaceanalysis}

As mentioned in \ref{sec:supernas}, the balance between model sparsity and capacity is essential in search space design, and the configurations of search space determine the supernet and single-path model architecture. 
Therefore, we need to evaluate search space before sampling supernet. 
Alg. \ref{alg:search_space} shows how we analyze the search space. 
Before the analysis, we set the hyper-parameter $\lambda_{lo}$ and $\lambda_{up}$ as the lower and upper bound of the single-path model's capacity. 
We randomly sample configurations $c_{sp}$, including sparsity and block-pruning configuration in each layer, from the search space and build sparse single-path models. 
If the sparse single-path model meets the hardware constraint, we evaluate the count of parameters in the model and accept the model if the count is between the given boundary. 
Finally, we calculate the statistical probability to represent how many sampled single-path models are acceptable in the search space. 
If the statistic probability ($n_{aceept}/n_{eval}$) is higher than $90\%$, we accept the search space for further deployment. 
Otherwise, we adjust the search space in advance to avoid unnecessary searches.
The adjustment of search space is performed by:

\begin{algorithm}[t]
\setstretch{1.1} 
\textbf{Input:} Search space $\mathbb{S}_{sup}$, memory limit ${\cal L}_{m}$ and storage limit ${\cal L}_{s}$, iteration count ${\cal T}_i$, ${\cal T}_j$.\\
\textbf{Output:} An optimal  supernet ${\cal A}_{sup}$.\\
\For {$i=1$ \textbf{to} ${\cal T}_i$}{
    $c_{sup}^i$ $\leftarrow$ $EvolutionarySample(\mathbb{S}_{sup})$ \\
    ${\cal A}^i$ $\leftarrow$ $CreateSupernet(c_{sup}^i)$\\
    
    \If {$TestSupernet({\cal A}^i)\neq True$}{
        \textbf{continue}
    }
    Train ${\cal A}^i$ for $10$ epochs\\
    \For {$j=1$ \textbf{to} ${\cal T}_j$}{
        $c_{spa}$ $\leftarrow$ $RandomSample(\mathbb{S}_{spa})$ \\
        ${M^{i,j}}$ $\leftarrow$ $CreateModel(c_{spa}, {\cal A}^i)$\\
        $[l_m,l_s]$ $\leftarrow$ $ResourceEval({\cal M}^{i,j},c_{sp}^k)$ \\
        \If {$l_m$ $>$ ${\cal L}_{m}$ \bf{or} $l_s$ $>$ ${\cal L}_{s}$}{
            \textbf{continue}
        }
        ${\cal M}_{prun}^{i,j,k}\leftarrow OneShotPrune({\cal M}^{i,j},c_{sp}^k)$\\
        $Acc \leftarrow AccuracyEval({\cal M}_{prun}^{i,j,k})$\\
        $AvgAcc^{i,j}\leftarrow  UpdateAvgAcc(Acc)$
        
        $AvgAcc^{i}\leftarrow  UpdateAvgAcc(AvgAcc^{i,j})$\\
    }
    ${\cal A}_{sup}\leftarrow Update Best({\cal A}^i,AvgAcc^{i})$\\
}
\textbf{return} ${\cal A}_{sup}$
\caption{Supernet Architecture Seach}
\label{alg:supernet_search}
\end{algorithm}

\begin{equation} \label{eq:1}
    y = Round(k(x-\overline{x})+b),
\end{equation}
where $x$ and $y$ are dimensional settings before- and after- adjustment. Symbols with overline refer to the average values. $Round$ functions projects floats to their nearest integers. The slope $k$ in equation \ref{eq:1} is given by:
\begin{equation} \label{eq:2}
    k = \frac{n_{accept}}{n_{eval}}+10\%,
\end{equation}
where $n_{accept}$ and $n_{eval}$ is counted in Alg. \ref{alg:search_space}. The intercept $b$ in equation \ref{eq:1} is given by:
\begin{equation} \label{eq:3}
    b=x_{min}+\frac{\overline{N_{params}}-N_{min}}{N_{max}-N_{min}}(x_{max}-x_{min}),
\end{equation}
where $N_{min}$ and $N_{max}$ are the minimum and maximum value in Parameters Evaluations ($N_{params}$).
$x_{min}$ and $x_{max}$ are named in the same way.

To be specific, only the dimensional setting will be adjusted, while the number of heads in MHSA and number of modules in \texttt{DoT} are fixed.
The search space can be also tuned manually to adapt the implementation of calculation (e.g set dimension to a multiple of 4).

In detailed implementations, we choose $\lambda_{lo}=0.8$ and $\lambda_{up}=2.8$ based on our preliminary experience.
\textcolor{black}{We conducted experiments in three types of search space: \textit{Small}, \textit{Normal}, and \textit{Large}, which denote the model size sampled from them.
The experiments results prove that the search space analysis allows SuperNAS to optimally balance sparsity and accuracy trade-offs.
Detailed experiments of search space analysis can be found in Sec. \ref{sec:offline}.}

\subsubsection{Supernet Architecture Search}
\label{subsubsec:supernet_search}

The search process of supernet is shown in Alg. \ref{alg:supernet_search}. 
For each sampled supernet, we perform a simple test on it before the actual search (expressed as \textit{TestSupernet} in Alg. \ref{alg:supernet_search}). 
Specifically, we randomly sample 100 single-path models from the supernet and evaluate the memory usage of each model. 
If half of the models exceed the memory limit of hardware constraint, we skip the search procedure for this supernet.

After the simple test of supernet, we take two steps to evaluate its sensitivity to sparsity.
Firstly, we randomly select single-path models from the supernet.
For each single-path model, compression is conducted by generated sparse configuration to check whether the model occupies more resources than the practical scenario.
Then we take the average accuracy as the performance metric of the supernet.
The configuration sampled in $\mathbb{S}_{sup}$ is updated with evolutionary algorithm (expressed as \textit{EvolutionarySample} in Alg. \ref{alg:supernet_search}).
The supernet with the highest performance metric will be adopted for the next search stage.

\begin{algorithm}[tb]
    \caption{Single-path Model Search in SparseNAS}
    \label{alg:single_search}
    \textbf{Input:} Search space $\mathbb{S}_{spa}$, supernet architecture ${\cal A}$, memory limit ${\cal L}_{m}$ and storage limit ${\cal L}_{s}$, iteration count ${\cal T}_{spa}$.\\
    \textbf{Output:} An optimal pruned single-path model ${\cal M}_{prun}$.\\
    \For {$i=1$ \textbf{to} ${\cal T}_{spa}$}{
        $c_{spa}$ $\leftarrow$ $EvolutionarySample(\mathbb{S}_{spa})$ \\
        ${M^{i}}$ $\leftarrow$ $CreateModel(c_{spa}, {\cal A})$\\
        $[l_m,l_s]$ $\leftarrow$ $ResourceEval({\cal M}^{i})$ \\
        \If {$l_m$ $>$ ${\cal L}_{m}$ \bf{or} $l_s$ $>$ ${\cal L}_{s}$}{
            \textbf{continue}
        }
        Train ${\cal M}^{i}$ for 10 epochs\\
        ${\cal M}_{prun}^{i}\leftarrow IterativePrune({\cal M}^{i})$\\
        $Acc^i\leftarrow AccuracyEval({\cal M}_{prun}^{i})$\\
        ${\cal M}_{prun}, c_{sp}\leftarrow UpdateBest({\cal M}_{prun}^{i},Acc^i)$\\
    }
    \textbf{Return} ${\cal M}_{prun}$
\end{algorithm}

To reduce the search cost, we evaluate the storage and peak memory usage of the sparse models for skipping the ones that do not satisfy the resource requirements.
In addition, the one-shot weight-magnitude pruning algorithm without tuning is adopted to reduce the runtime cost (presented as \textit{OneShotPrune} in Alg. \ref{alg:supernet_search}).

\subsection{SparseNAS: Hardware-Aware Sparse Model Search} \label{sec:sparsenas}

SparseNAS aims to obtain the best sparse single-path model from the supernet. 
Differently from the original SPOS approaches from \cite{guo2020single}, the compression procedure is performed, including pruning and quantization, among the searching steps.
Moreover, in order to reduce the cost of training and compression procedures, SparseNAS is divided into two stages.
In first stage (selection stage), SparseNAS aims to find the best sparse single-path model with \textit{pruning} and \textit{quantization} procedures.
For the second stage (compression stage), SparseNAS only performs pruning and fine-tuning operations to recover the model's accuracy.

\begin{figure*}[t]
    \centering
    \begin{minipage}[t]{1.0\columnwidth}
        \centering
        \includegraphics[width=\columnwidth]{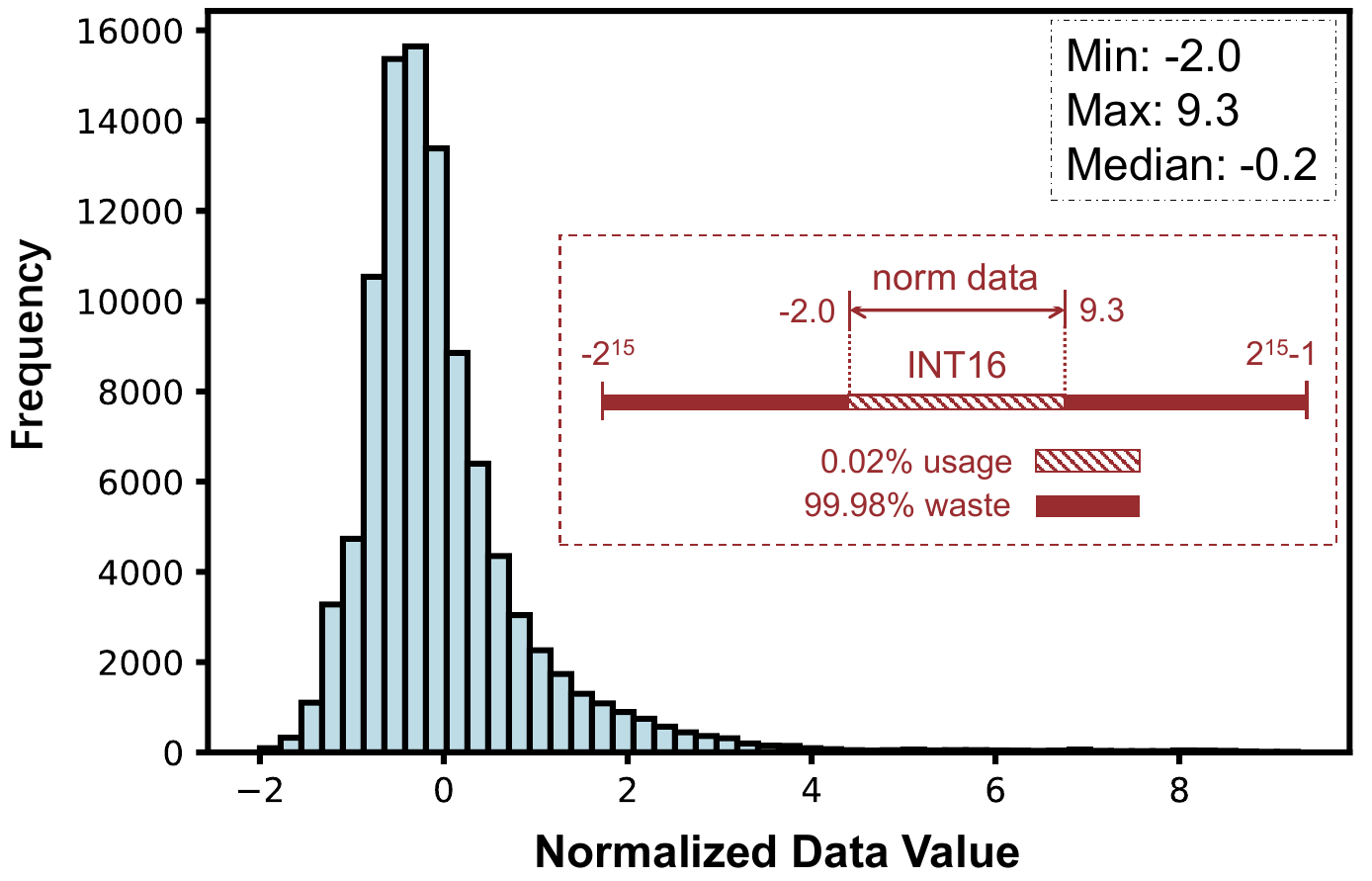}
        \caption*{(a)}
    \end{minipage}
    \hspace{0.1\columnwidth}
    \begin{minipage}[t]{0.6\columnwidth}
        \centering
        \includegraphics[width=\columnwidth]{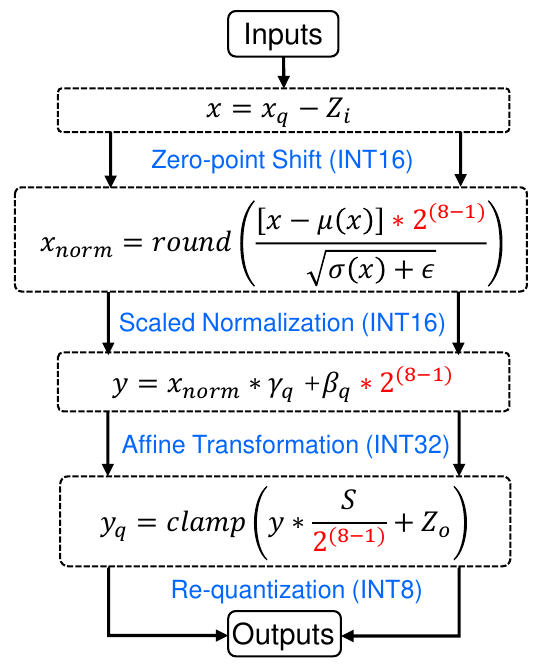}
        \caption*{(b)}
    \end{minipage}
    \vspace{-4pt}
    \caption{(a) Distribution of normalized activation in the first LayerNorm layer. (b) \textit{Scaled-LayerNorm} inference process. $S$ and $Z$ are the scaled and zero-point parameter of quantization respectively. $clamp$ function clamps the result to range of signed \texttt{INT8}.}
    \label{fig:fig-tinyformer-distribution_qlayernorm}
\end{figure*}

\begin{figure}[t]
    \centering 
    \includegraphics[width=1.0\columnwidth]{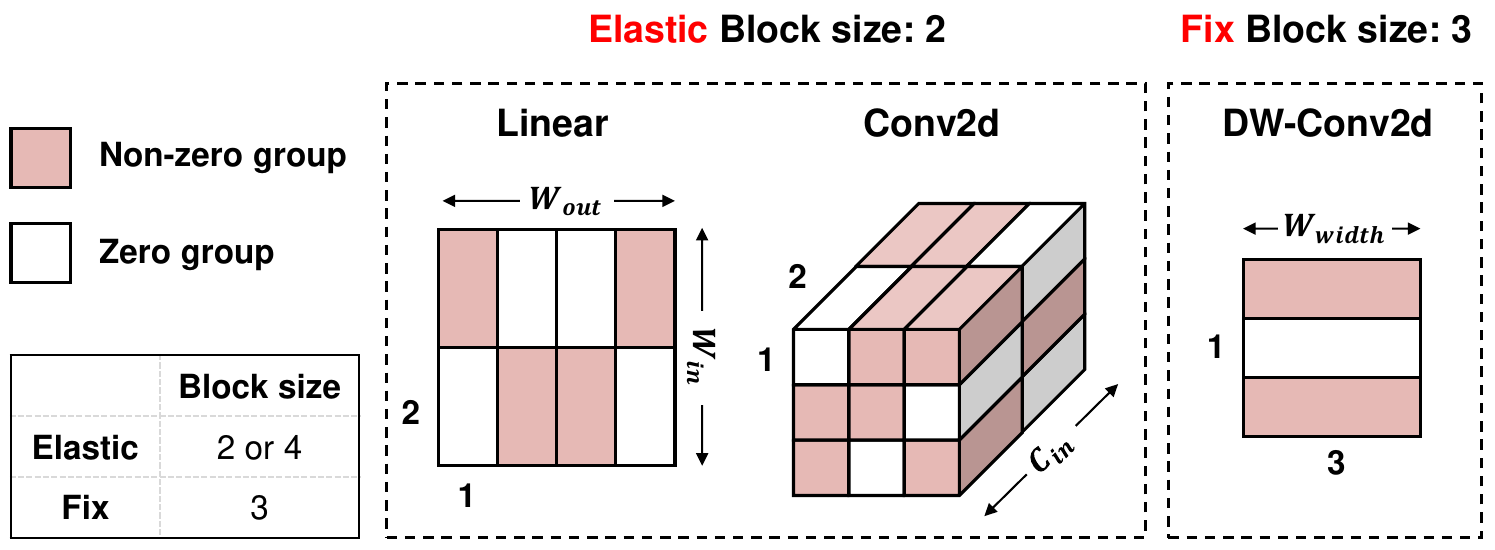}
    \caption{Blockwise pruning method.
    We abbreviate depthwise convolution in 2d as \texttt{DW-Conv2d}.
    $W_{in}$ and $W_{out}$ denote the dimensions of linear layer, $C_{in}$ means the number of input channels and $W_{width}$ is the kernel width of depthwise convolution.
    }
    \label{fig:fig-tinyformer-prune}
\end{figure}

\subsubsection{Two-stage process}

SparseNAS consists of the selection stage and the compression stage. 
The selection stage of SparseNAS is presented in Alg. \ref{alg:single_search} for single-path model selection.
In the selection stage, SparseNAS trains randomly-sampled sparse single-path models for a few epoch, then performing iterative pruning and accuracy evaluation. 
The model with highest accuracy is sent to the compression stage for further training.
We skip the model candidates that do not satisfy hardware constraints by evaluating the storage and memory usage.

For the compression stage, SparseNAS performs iterative pruning again and fine-tuning on the model for accuracy improvement. 
After the compression stage, the model will be sent to SparseEngine for efficient deployment. 
Two-stage procedures reduce the training and compression cost while maintaining the obtained single-path model's accuracy.

\subsubsection{Pruning Method}
Different from the one-shot pruning method in SuperNAS, AGP iterative pruning method \cite{Zhu2018ToPO} is utilized in two-stage NAS.
AGP method can avoid significant accuracy degradation caused by pruning.
We only perform weight pruning in \texttt{Conv2d} and \texttt{Linear} operators.

SparseNAS utilizes a blockwise pruning method, grouping multiple continuous weights as a block to prune.
Pruning configuration (sparsity and block size) affects both accuracy and hardware resource usage in deployment, and the effects vary from different layers.
Therefore, we adopt a mixed-blockwise pruning strategy in \texttt{Conv2d} and \texttt{Linear} layers, as shown in Fig. \ref{fig:fig-tinyformer-prune}.
In mixed-blockwise pruning procedure, SparseNAS selects elastic block size (2 or 4) for each layer to prune.
When sampling the single-path model, each choice module is set to a random configuration of sparsity and block size.

In SparseEngine, we applied the blockwise convolution in width direction to exploit spatial locality in computation.
Therefore, as an exception, the block size of blockwise convolution is set to $3$, for the kernel size is fixed to $3$$\times$$3$.

\subsubsection{Quantization Method}
\label{sub:quant}
Floating-point calculations on MCUs are inferior in latency and power consumption compared to integer calculations.
Therefore, we quantize the model weights and activations to \texttt{INT8} by Post-Training Quantization (PTQ) algorithm \cite{krishnamoorthi2018quant}.
However, LayerNorm calculations in transformer is not suitable for directly quantized. 
Performing linear quantization on LayerNorm will cause a significant accuracy drop.
The original LayerNorm is defined as:
\begin{equation}
    y=\frac{x-\mu(x)}{\sqrt{\sigma(x)+\epsilon}}*\gamma+\beta,
\end{equation}
where $\mu(x)=\frac{1}{c}\Sigma^c_{i=1}x_i$ and $\sigma(x)=\sqrt{\frac{1}{c}\Sigma^c_{i=1}(x_i-\mu)^2}$ are the mean and variance values of input $x$ in channel-wise direction ($c$ is the number of channels).  
$x_{norm}=\frac{x-\mu(x)}{\sqrt{\sigma(x)+\epsilon}}$ is the normalization result before affine transformation.
$\gamma$ and $\beta$ are the learnable parameters in affine transformation. $\epsilon$ is a significantly small value to prevent the denominator to be zero.
In linear quantization, the normalization result $x_{norm}$ is supposed to be rounded to integer.
However, directly rounding $x_{norm}$ to \texttt{INT16} incurs a significant loss of precision.

To verify the conjecture, we count the distribution of $x_{norm}$ of LayerNorm in first \texttt{DoT} Architecture.
As shown in Fig. \ref{fig:fig-tinyformer-distribution_qlayernorm}(a), the normalized data is mapped to the range of $-2.0$ to $9.3$.
Rounding the small range of normalization results to \texttt{INT16} causes a large loss of precision.
On the other hand, the range of \texttt{INT16} is not fully utilized by $x_{norm}$.
Adopting \texttt{INT16} to store the normalized data will waste $99.98\%$ of integer values, presented in Fig. \ref{fig:fig-tinyformer-distribution_qlayernorm}(a).

To tackle this problem, \textit{Scaled-LayerNorm} is proposed to perform integer-only inference instead of naive quantized LayerNorm.
As shown in Fig. \ref{fig:fig-tinyformer-distribution_qlayernorm}(b), we enlarge the normalized results by $2^{(8-1)}$ to reduce the numeric precision loss in quantization.
After the linear transformation, the results are stored as \texttt{INT16} format, while it also includes the $2^{(8-1)}$ factor.
In re-quantization step, we fold $\frac{1}{2^{(8-1)}}$ into scaling factor $S$ to ensure mathematical equivalence.
Expanding the numerical range of $x_{norm}$ prevents significant accuracy drop caused by large precision loss.
With the approaches above, we reach a significant speed-up in LayerNorm inference, with only slight accuracy drop.
Detailed experimental results of Scaled-LayerNorm is presented in Sec. \ref{sub-runtime}.

\subsection{SparseEngine: Efficient Deployment Library of Sparse Transformers} \label{sec:sparseengine}

\begin{figure*}[t]
    \centering
    \includegraphics[scale=0.5]{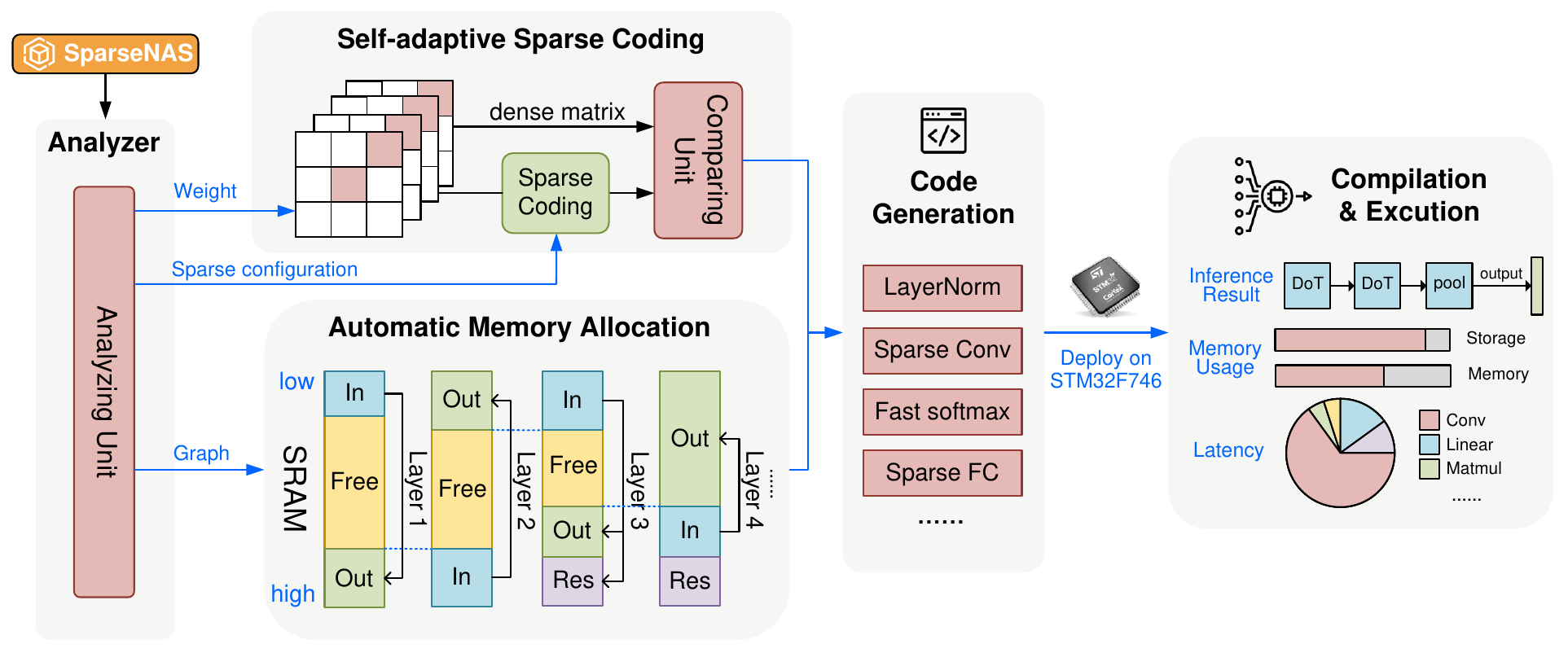}
    \caption{The workflow overview of SparseEngine.
    Sparse models obtained from SparseNAS are first analyzed for sparse coding and memory allocation. Then, different kinds of layers' computations are transformed via automatic code generation for the following compilation and execution on targeted MCUs.}
    \label{fig:fig-tinyformer-sparseengine}
\end{figure*}

SparseEngine is a deployment tool for sparse transformers on MCUs.
It consists of a deployment library based on C++ language, and a code generator.
SparseEngine can automatically allocate memory to each layer and generate codes that can directly deploy on STM32 MCUs.
Compared with CMSIS-NN and TinyEngine, SparseEngine supports extra functions, including dynamic sparse calculate of \texttt{Conv2d} and \texttt{Linear} operations, to efficiently inference a sparse transformer. 
Moreover, SparseEngine optimizes \textit{softmax} operation, which is more frequently used in transformer inference.
Comprehensive details of SparseEngine are illustrated in Fig. \ref{fig:fig-tinyformer-sparseengine}.

Firstly, the models obtained from SparseNAS are analyzed to extract the required information, including sparse configuration, model architecture and memory usage, etc.
To efficiently utilize the available memory on MCUs, a head-tail-alternation allocation strategy is adopted to automatically allocate memory for models inference, presented in Fig. \ref{fig:fig-tinyformer-sparseengine}.
Meanwhile, inspired by the run-length coding algorithm \cite{Robinson67runlength}, we perform the self-adaptive sparse strategy with blockwise run-length coding in \texttt{UINT8}.
With self-adaptive strategy, weights are stored as sparse format only if sparse coding could reduce the storage occupation.
\textcolor{black}{Finally, targeted codes could be generated for deployment on MCUs.
The code generator in SparseEngine converts ONNX model to C-like code executable on on MCU.
The adoption of ONNX as an intermediate representation enables TinyFormer to maintain interoperability across multiple machine learning pipelines.}

Compared with other deployment approaches on MCUs, SparseEngine aims to further exploit the sparsity on TinyML deployment and model inference.
The sparsity exploitation includes sparse encoding/decoding, and sparse calculations (both \texttt{Conv2d} and \texttt{Linear} layers). 
Moreover, targeting on model inference with transformer modules, \texttt{Softmax} operator is also optimized to accelerate the inference on MCUs.

\begin{figure}[t]
    \centering
    \includegraphics[width=1.0\columnwidth]{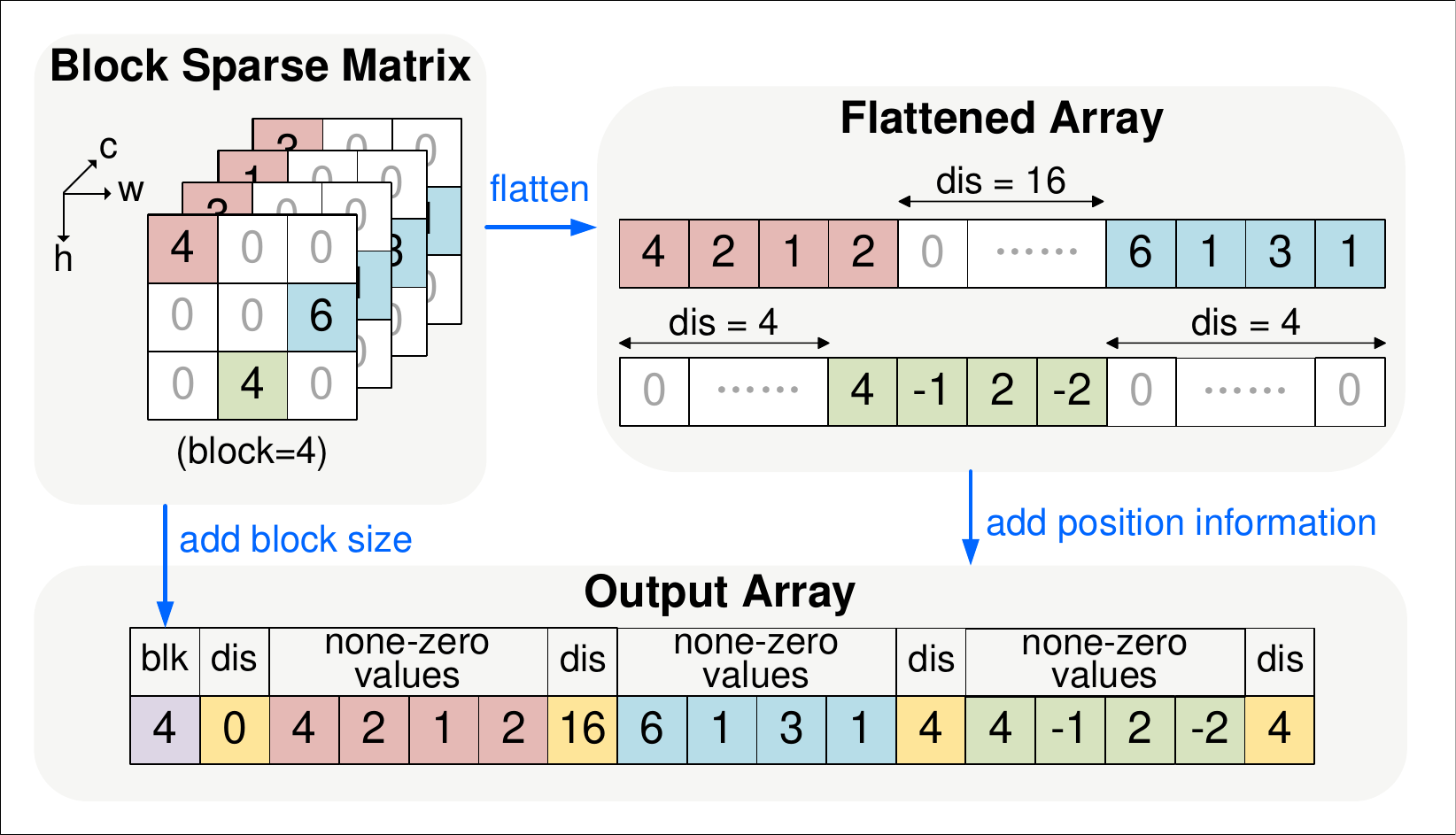}
    \caption{Blockwise run-length coding for 3D matrix.}
    \label{fig:fig-sparse-coding}
\end{figure}

\subsubsection{Sparse Coding}
\label{subsubsection-sparsecoding}

In sparse coding, based on run-length coding algorithm \cite{Robinson67runlength}, we perform blockwise run-length coding in 8-bit to adopt block pruning.
At first. the original 3D matrix (tensor) is flattened into array format. 
The spare weights are stored in $(dis, val_1, val_2 ...,val_b)$ format, as presented in Fig.~\ref{fig:fig-sparse-coding}, where $dis$ represents the distance between non-zero blocks as the position information, $val_i$ indicates the $i$-th weight in the non-zero block whose length is $b$.
We insert zero elements if $dis$ is beyond the maximum value of \texttt{INT8}.

Compared with Coordinate (COO) and Compressed Sparse Row (CSR) formats, blockwise run-length coding has a higher compression ratio. 
It only requires one element to represent the position of adjacent non-zero weights. 
The compression ratio of this encoding format could be obtained by
\begin{equation}
    \eta =\frac{1}{( 1-\rho)\times (1+\frac{1}{b})},
\end{equation}
where $\eta$ indicates the compression ratio, $\rho$ refers to the sparsity and $b$ is the block size of pruning. 
Consequently, the compression ratio of blockwise run-length encoding could be larger along with the increasing of sparsity and block size.
Cooperating with mixed-blockwise pruning in SparseNAS, blockwise run-length encoding significantly reduce the required size of sparse coding.

\subsubsection{Sparse Convolution and Linear}

\begin{figure}[t]
    \centering
    \includegraphics[width=1.0\columnwidth]{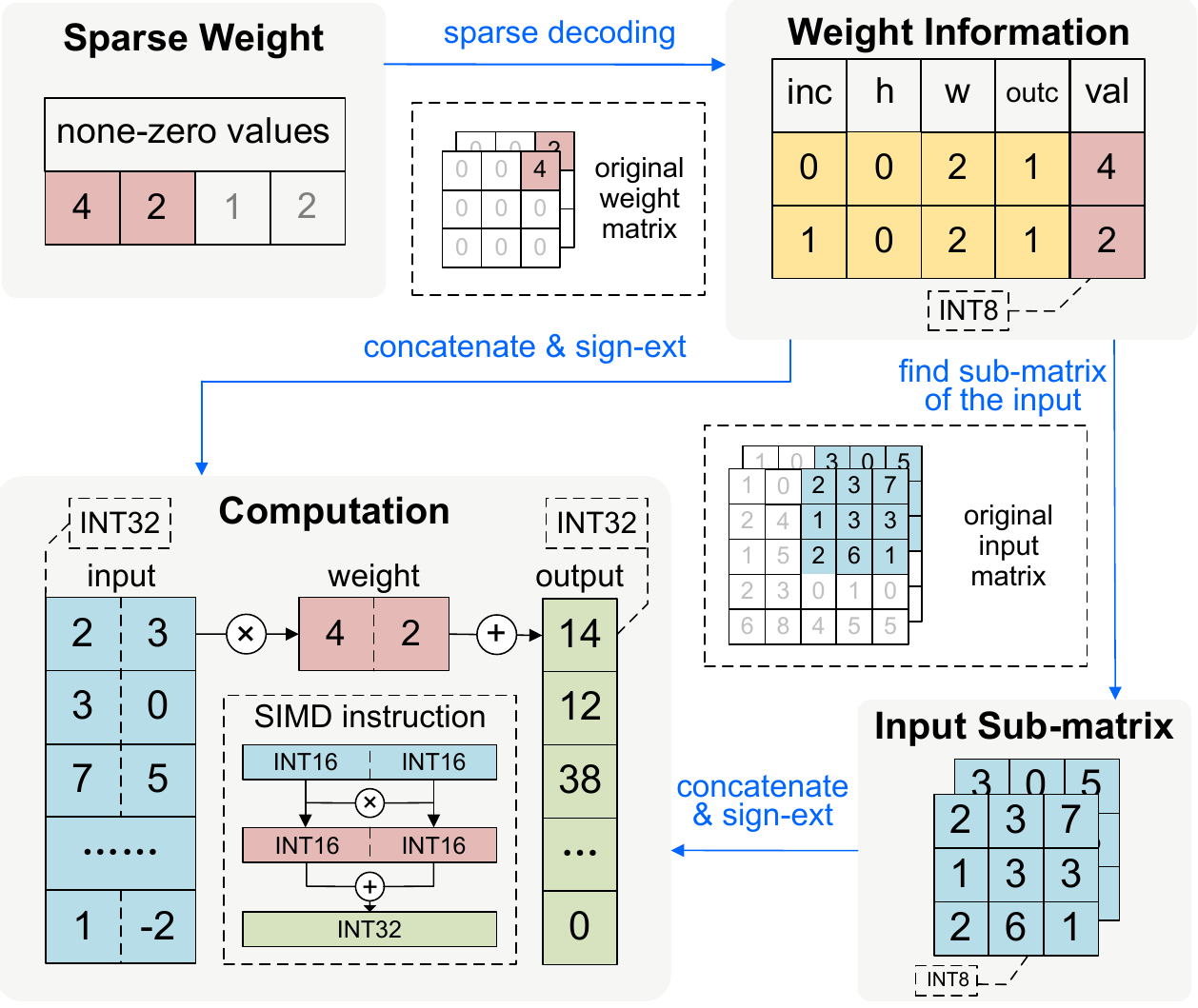}
    \caption{Sparse Convolution with SIMD instruction.
    Each block's weight information are extracted by sparse decoding procedure.
    Among them, the location information are adopted to find the sub-matrix in original input matrix.
    After concatenate \& sign extension 8(sign-ext) operations, sub-matrix and weight values are sent for computations with SIMD instructions acceleration.}
    \label{fig:fig-tinyformer-sparseconv}
\end{figure}

Decoding sparse weights and then computing convolution in dense format occupies a large amount of memory footprint \cite{yang2021s, dai2020sparsetrain}.
To avoid unnecessary memory usage, we perform sparse convolution calculation directly in sparse format. 
Fig. \ref{fig:fig-tinyformer-sparseconv} presents the details of the sparse \texttt{Conv} calculation.
Sparse weights are decoded to obtain the coordinates and values.
Each weight value corresponds to a sub-matrix of the input matrix.
After extracting the sub-matrix, we execute element-matrix multiplication and accumulate the results as the output. 
Specifically, we decode two weight values simultaneously and extract two corresponding elements from the sub-matrix. 
Then, two \texttt{INT8} values are sign-extended to \texttt{INT16} and concatenated as \texttt{INT32} format. 
Thus, the above multiply-accumulate calculations can be performed by SIMD instructions.
Meanwhile, the sparse linear layers are implemented by the same manner, despite the difference in dimensions.



\subsubsection{Softmax Optimization}
According to our profiling, the \texttt{Softmax} layer is very time-consuming due to the required exponential operations.
The computation of \texttt{Softmax} can be described as
\begin{equation}
    Softmax(x_i)=\frac{e^{x_i}}{\sum_{j=1}^n e^{x_j}}=\frac{e^{x_i-x_{max}}}{\sum_{j=1}^n e^{x_j-x_{max}}},
\end{equation}
where $x_i$ indicates the $i$-th value and $x_{max}$ is the maximum value in $x$ ($n$ is the number of elements in one channel).
Since the inputs of \texttt{Softmax} are in \texttt{INT8} format, the exponents range from -$256$ to $0$.
Consequently, redundant calculations will be increased according to the input size of \texttt{Softmax}.
Therefore, we optimize the \texttt{Softmax} operator with a lookup table to reduce redundant computations.

Targeting at calculating the negative exponential functions, SparseEngine adopts the absolute value of the exponential factor as an index to query the bitmap.
If the corresponding bit exists in the bitmap, SparseEngine obtains the result from the lookup table and reuses it in \texttt{Softmax} computations.
Otherwise, it calculates the exponential function and stores the result in the table, updating the corresponding bit in the bitmap.
According to the evaluation of SparseEngine, its memory usage on bitmap and lookup table is less than $1.2$KB.
Since SparseEngine reuses the buffer memory of convolution when performing \texttt{Softmax} optimization, there is no extra memory cost in calculations.

\begin{table*}[t]
\renewcommand\arraystretch{1.25}
\setlength\tabcolsep{8pt}
\caption{Top-1 accuracy comparison and hardware resources evaluation.
Different models are evaluated under different hardware constraints.
TinyFormer-120K and TinyFormer-300K indicate that the peak memory limit is set as $120$KB and $300$KB, respectively.
Texts that end with $\ast$ indicate that the peak memory or storage usage is beyond the limitations of STM32F746 MCU.}
\label{tab:summary}
\begin{center}
\scalebox{1.0}{
\begin{tabular}{l|ccccc}
\hline
Model & Peak Mem. & Storage & CIFAR-10 & ImageNet-32\\
\hline
MobileNetV2 \cite{Sandler2018MobileNetV2IR} & $416$KB$\ast$  & $2.13$MB$\ast$ & $94.61$\%  & $38.22\%$\\
CCT-7/3$\times$1 \cite{hassani2021escaping} & $512$KB$\ast$  & $3.52$MB$\ast$ & $95.72$\% & $39.04$\% \\
MobileViT-XS \cite{mehta2021mobilevit} & $160$KB  & $1.91$MB$\ast$ & $90.11$\%  & $34.16\%$\\
MobileViTV2-0.75 \cite{MehtaR23} & $172$KB  & $3.48$MB$\ast$ & $91.30\%$  & $34.99\%$\\
MCUNet-in3 \cite{lin2020mcunet, lin2021mcunetv2} & $22$KB  & $0.89$MB & $84.26\%$  & $26.60\%$\\
\hline
TinyFormer-120K (ours)  & $120$KB  & $0.94$MB  & $94.52$\% & $38.53$\%\\
TinyFormer-300K (ours) & $300$KB  & $0.91$MB  & $\boldsymbol{96.10\%}$ & $\boldsymbol{39.37\%}$\\
\hline
\end{tabular}}
\end{center}
\end{table*}

%% file: 4-experiment.tex
\section{Experimental Results} \label{sec:experiment}
\textcolor{black}{Here we present experimental results to demonstrate the effectiveness of our TinyFormer framework.
The results mainly consist of offline evaluation and runtime validation.
In Sec. \ref{sec:offline}, we provide offline evaluation results with algorithmic performance to demonstrate the effectiveness of SuperNAS and SparseNAS.
In Sec. \ref{sub-runtime}, we are focused on the runtime efficiency on MCU platform, and present the runtime validation of SparseEngine.}

\subsection{Evaluation Setup}

\subsubsection{Dataset}
According to requirements in TinyML scenario, our TinyFormer is mainly evaluated on CIFAR-10 \cite{Krizhevsky2009LearningML} and ImageNet-32 \cite{ChrabaszczLH17imagenet32} dataset for image classiﬁcation.
CIFAR-10 contains 50 thousand and 10 thousand images for training and validation respectively.
ImageNet-32 contains 1.28 million and 50 thousand images for training and validation respectively.
Images in both datasets are set at $32\times32$ resolution in RGB format.

\subsubsection{Training Settings}
\label{sub:trainingset}
In the training process, we use AdamW \cite{Loshchilov2019DecoupledWD} as the optimizer.
The learning rate starts with a warm-up phase, increasing from $1\times10^{-6}$ to $5.5\times10^{-4}$ for the first $10$ epochs, and then follows a cosine annealing schedule, gradually decreasing to $1\times10^{-5}$.
During the model's training, label smoothing with a probability of $0.1$ is employed, as suggested by Szegedy et al. \cite{SzegedyVISW16}.
Unless otherwise specified, all involved models are trained from scratch for $300$ epochs using a batch size of $128$.

The weights of models are initialized using the method from \cite{HeZRS15}.
All models are performed \texttt{INT8} quantization.
The reported top-1 accuracy of image classification task is evaluated on full-integer models if not specifically stated.

Some baseline models are designed towards the ImageNet dataset with higher resolution. 
To ensure a fair comparison, we maintain the original architecture of baseline models, except for the number of classes in classification.

\subsubsection{Platforms}
\textcolor{black}{The offline experiments in Sec. \ref{sec:offline} are conducted on 8 NVIDIA V100 GPU.
The runtime experiments in Sec. \ref{sub-runtime} are conducted on with SparseEngine for accuracy and efficiency evaluation. 
We select STM32F746 MCU platform (Cortex-M7 core @ 216MHz, with 320KB RAM and 1MB Flash) and STM32H743 MCU platform (Cortex-M7 core @ 480MHz, with 512KB RAM and 2MB Flash) to conduct the runtime experiments.
All program is burned by Keil5 MDK supporting ARM-based microcontrollers.}


\subsection{Offline Evaluation} \label{sec:offline}

In offline evaluation, we have mainly trained two model with different hardware constraints (TinyFormer-300K and TinyFormer-120K). In the NAS process, SuperNAS takes 15.2 GPU hours per sampling, and SparseNAS takes 1.5 GPU hours per sampling.

Tab. \ref{tab:summary} shows the comparison results of our TinyFormer and other state-of-the-art lightweight models.
With the co-optimization of SuperNAS and SparseNAS, TinyFormer-300K could satisfy the hardware constraints of STM32F746 MCU platform and achieve a record accuracy with CIFAR-10 and ImageNet-32 on MCUs. 
As illustrated in Fig. \ref{fig:fig-tinyformer-supernet}, TinyFormer is designed as a hybrid model that contains both convolution and transformer encoder layers. 
Compared to other lightweight hybrid models, such as MobileViT-XS, TinyFormer better combines the advantages of CNN and transformer to achieve higher accuracy with limited resources. 
Compared with CCT-7/3$\times$1 designed for CIFAR-10, TinyFormer-300K achieves higher accuracy while reducing peak memory and storage by $41\%$ and $74\%$, respectively.
Meanwhile, TinyFormer-120K is searched for stricter peak memory constraints. 
Compared with MobileViT-XS, TinyFormer-120K improves the accuracy by $4.4\%$ with less peak memory and storage.

\subsubsection{Search Space}

\begin{figure*}[t]
    \centering
    \begin{minipage}[t]{0.8\columnwidth}
        \centering
        \includegraphics[width=\columnwidth]{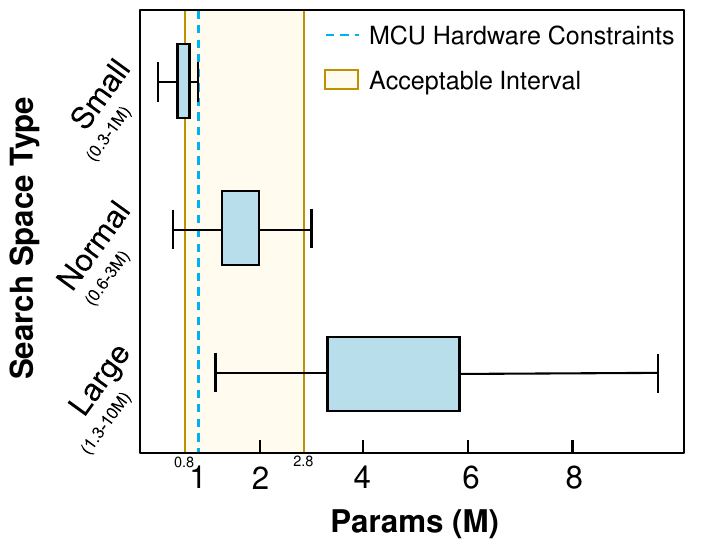}
        \caption*{(a)}
    \end{minipage}
    \hspace{0.1\columnwidth}
    \begin{minipage}[t]{0.8\columnwidth}
        \centering
        \includegraphics[width=\columnwidth]{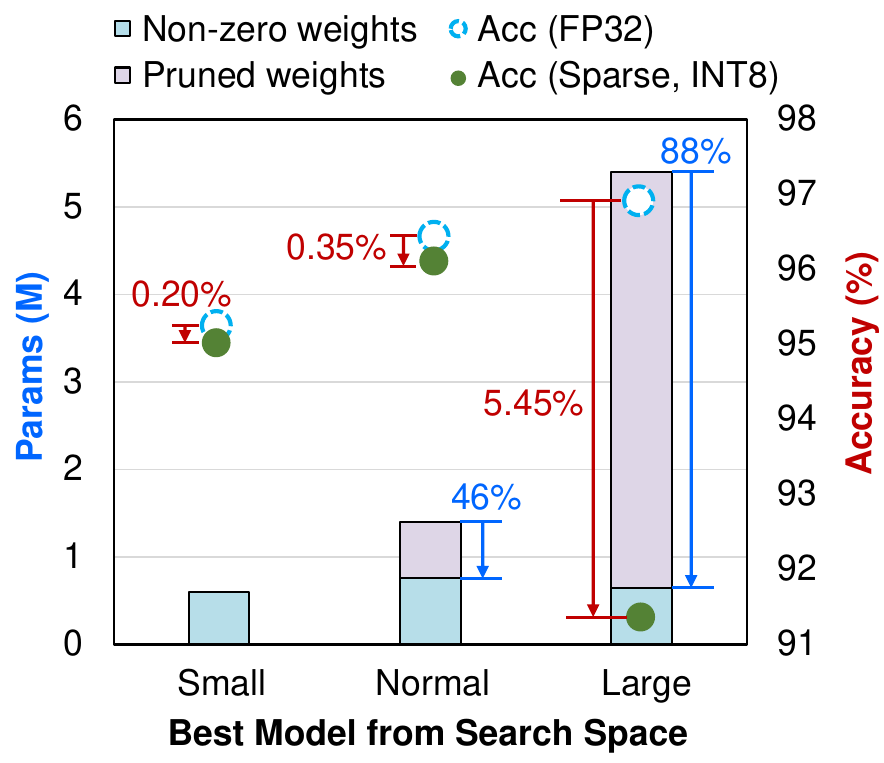}
        \caption*{(b)}
    \end{minipage}
    \vspace{-4pt}
    \caption{(a) The obtained model size from three types of search space.
    The model sampled from a small search space satisfies the hardware constraints without pruning.
    The models sampled from large search spaces require compression to meet the constraints.
    (b) The best-performing model in each search space before/after the quantization and pruning process.
    }
    \label{fig:offline-searchspace}
\end{figure*}

\begin{table}[t]
\renewcommand\arraystretch{1.25}
\caption{Ablation of TinyFormer in offline evaluation. Basic TinyFormer structure has two \texttt{DoT}s and mixed block size. 
All the models are searched with hardware constraints of $300$KB Memory and $980$KB Storage in CIFAR-10. \#Params indicate the number of effective weights.}
\label{tab:ablation}
\begin{center}
\scalebox{0.9}{
\begin{tabular}{l|ccc}
\hline
Model & Accuracy & \#Params   & Storage \\
\hline
TinyFormer   & $\boldsymbol{96.10\%}$   & $731$K  & $942$KB\\
TinyFormer (w/o Tr.)   & $94.62\%$   & $756$K  & $963$KB\\
TinyFormer (Single \texttt{DoT})   & $93.92\%$   & $572$K  & $655$KB\\
TinyFormer (Block Size $=2$)       & $95.88\%$   & $730$K  & $950$KB\\
TinyFormer (Block Size $=4$)       & $95.52\%$   & $807$K  & $962$KB\\
\hline
\end{tabular}}
\end{center}
\end{table}

The search space design affects the sampled model size.
An inappropriate search space have a negative impact on the experimental results.
As shown in Fig. \ref{fig:offline-searchspace}(a), we conduct experiments on three different sizes of search spaces based on the parameters counts in sampled supernet, including \textit{small} (0.3$\sim$1M), \textit{normal} (0.6$\sim$3M), and \textit{large} (1.3$\sim$10M).
With hyper-parameter $\lambda_{lo}=0.8$ and $\lambda_{hi}=2.8$, only \textit{normal} space has acceptance probability higher than $90\%$.
The \textit{small} and \textit{large} space has an acceptance probability less than $30\%$.
The \textit{small} search space mostly contains dense models with low capacity, while models in  \textit{large} search space have both higher parameter counts and higher sparsity.
\textcolor{black}{Figure \ref{fig:offline-searchspace}(b) illustrates trade-off between sparsity and accuracy. 
We set three search space: \textit{Small}, \textit{Normal}, and \textit{Large}.
The models in \textit{Small} search space are with lower scale and lower sparsity, 
The \textit{Large} search space demonstrates the inverse relationship.
The \textit{Normal} search space represents an intermediate configuration.
Within the \textit{Small} search space, lower sparsity results in lower accuracy degradation (0.20\%). 
However, the accuracy of model in \textit{Small} search space is substantially constrained by the hardware constraints.
By contrast, the \textit{Large} search space searches dense models with highest accuracy, but experiences more significant accuracy degradation (5.45\%) during pruning for MCU deployment.
The \textit{Normal} search space optimally balances sparsity and accuracy trade-offs, with higher accuracy of dense model compared with small search space, and less accuracy degradation during pruning compared with large search space.}

\subsubsection{Ablation Study}

\begin{figure}[t]
    \centering
    \includegraphics[width=1.0\columnwidth]{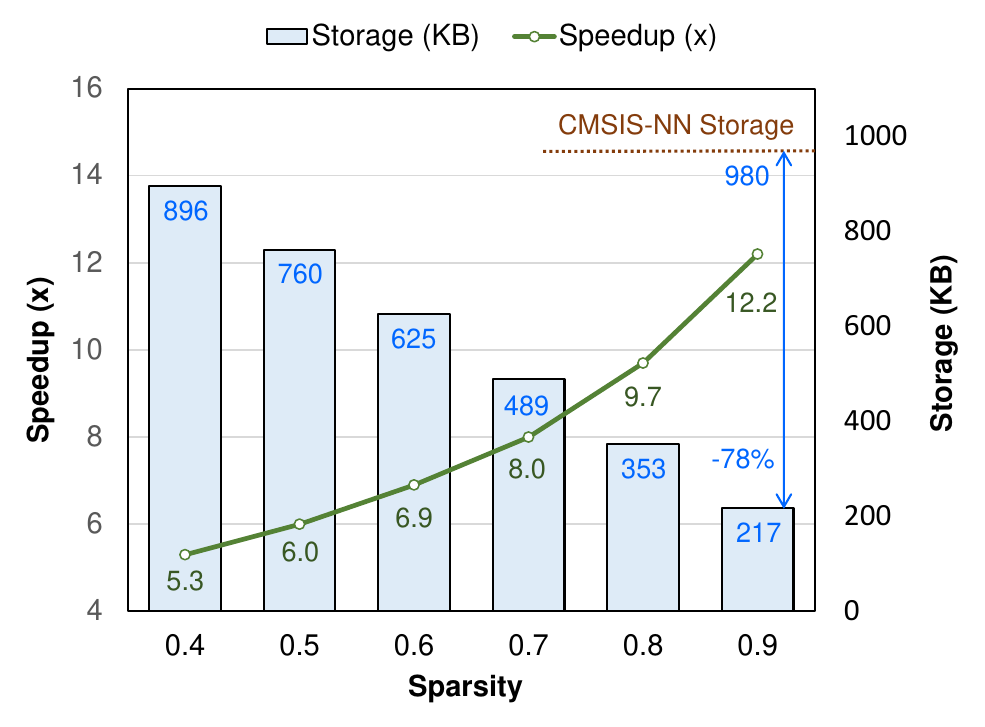}
    \caption{Performance comparison of SparseEngine and CMSIS-NN on CIFAR-10. SparseEngine is $5.3\times$$\sim$$12.2\times$ faster than CMSIS-NN under different sparsity, while saving $9\%$$\sim$$78\%$ storage.}
    \label{fig:fig-runtime-sparsity}
\end{figure}

In this part, we show the ablation study on TinyFormer from Tab. \ref{tab:ablation}, which consists of three perspectives: whether to use \textit{transformer block}, number of \texttt{DoT}s, and whether to adopt mixed block size pruning in search stage.
In Tab. \ref{tab:ablation}, TinyFormer refers to the TinyFormer-300K model in Tab. \ref{tab:summary}, with \textit{transformer block}, two \texttt{DoT}s, and mixed block size pruning.

To discover the impact of \textit{transformer block} in TinyFormer, a CNN model without \textit{transformer block} is searched and denoted as TinyFormer (w/o Tr.).
TinyFormer (w/o Tr.) deletes the \textit{transformer block} in \texttt{DoT} architecture, and only remains the \textit{downsample block} and \textit{MobileNetV2 block}.
Compared to TinyFormer (w/o Tr.), TinyFormer obtains better accuracy under almost the same resource usage.
These results suggest that incorporating transformer-related blocks or layers can offer advantages in achieving improved performance for TinyML.

Additionally, we conduct experiments to evaluate the impact of the number of \texttt{DoT} layers on the model's accuracy.
TinyFormer (Single \texttt{DoT}) is derived from the supernet that exclusively consists of a single \texttt{DoT} architecture, while adhering to the same hardware constraints.
With the same search space, TinyFormer (Single \texttt{DoT}) utilizes only $66.8\%$ of the storage limitation, resulting in a decrease in accuracy by $1.48\%$.
When employing a single \texttt{DoT}, the main bottleneck for TinyFormer (Single \texttt{DoT}) arises from memory constraints.
By contrast, the model with two \texttt{DoT}s almost approaches the limits of both storage and memory, effectively maximizing resource utilization.
Therefore, two \texttt{DoT}s are utilized in our basic experiments.

Finally, we evaluate the effectiveness of the mixed block size strategy in two stages of SparseNAS.
In particular, TinyFormer (Block Size $=2$) and TinyFormer (Block Size $=4$) denote models with a fixed block size of $2$ and $4$ in block pruning respectively.
Tab. \ref{tab:ablation} demonstrates that larger block sizes allow for more efficient storage of weights within the given limitations.
However, setting all block sizes to $4$ adversely affects the model's accuracy.
Based on these results, the pruning method employing a mixed block size scheme strikes the best balance between block size and the number of effective weights, thereby yielding the most suitable model with optimal performance.

\begin{figure}[t]
    \centering
    \begin{minipage}[t]{0.55\columnwidth}
        \centering
        \includegraphics[width=\columnwidth]{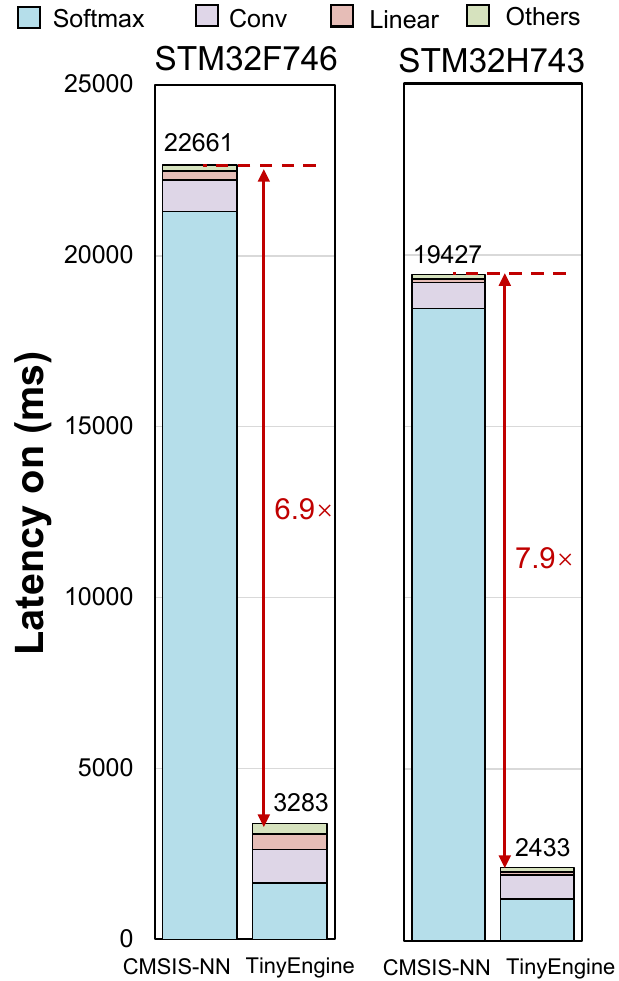}
        \caption*{(a)}
    \end{minipage}
    \hspace{7pt}
    \begin{minipage}[t]{0.38\columnwidth}
        \centering
        \includegraphics[width=\columnwidth]{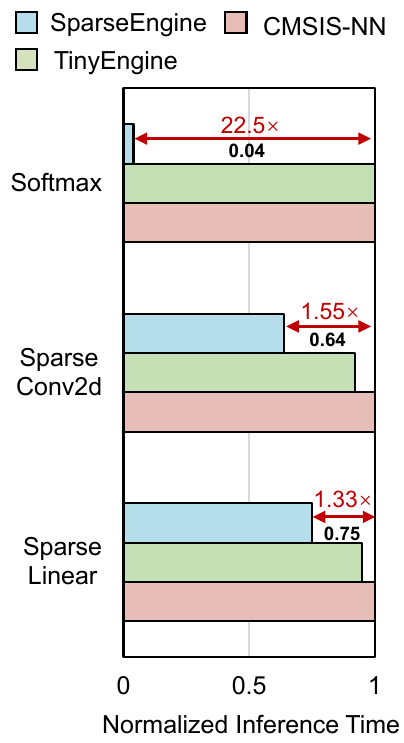}
        \caption*{(b)}
    \end{minipage}
    \caption{\textcolor{black}{(a) In CMSIS-NN, \texttt{Softmax} operator occupies $94\%$ of the latency. SparseEngine optimizes the procedure of texttt{Softmax} to achieve $6.9\times$ acceleration on STM32F746 and $7.9\times$ acceleration on STM32H43, with the average of $60\%$ sparsity. (b) Normalized inference time of SparseEngine, TinyEngine and CMSIS-NN on difference operators.}}
    \label{fig:runtime-latency}
\end{figure}

\subsection{Runtime Evaluation}
\label{sub-runtime}

At the runtime level, we have developed SparseEngine for performing sparse inference on MCUs.
The implementation of SparseEngine has successfully reduced the inference time to $3.8$ seconds on our highest-accuracy searched model.
To evaluate the performance of SparseEngine, we deploy the same sparse model in both CMSIS-NN and SparseEngine.
As illustrated in Fig. \ref{fig:fig-runtime-sparsity}, SparseEngine outperforms CMSIS-NN both in terms of inference latency and storage occupation.
By leveraging sparse inference support, SparseEngine achieves a significant reduction in storage requirements on MCUs, ranging from $9\%$ to $78\%$.
Through the utilization of specialized optimizations, SparseEngine achieves an impressive acceleration of inference, ranging from $5.3\times$ to $12.2\times$.
Specifically, in Fig. \ref{fig:fig-runtime-sparsity}, we mainly focus on how sparse configurations affect the inference speed and storage usage, rather than the accuracy.
For the simplicity of presentation, we have not shown the accuracy of models in the figure.

\begin{table}[t]
\renewcommand\arraystretch{1.25}
\caption{Comparasions of LayerNorm (LN) inference.
Experimental configurations are similar to TinyFormer-300K in Tab. \ref{tab:summary} except for the LayerNorm implementations.
We make an ablation study on different methods of LN: inference with \texttt{FP32} format (\texttt{FP32}-LN), inference with naive quantization to \texttt{INT8} (\texttt{INT8}-LN (naive)), and our proposed method in Sec. \ref{sub:quant} (\texttt{INT8}-LN (scaled))}
\label{tab:layernorm}
\begin{center}
\scalebox{1.0}{
\begin{tabular}{l|c|c|c}
\hline
Method of LN & Accuracy & Shape & Latency \\
\hline
\multirow{2}{*}{\texttt{FP32}-LN} & \multirow{2}{*}{$96.11\%$} & $[256\times44]$ & $194$ms\\
\cline{3-4}
 & & $[64\times192]$ & $208$ms \\
\hline
 \multirow{2}{*}{\texttt{INT8}-LN (naive)} & \multirow{2}{*}{$95.92\%$} & $[256\times44]$ & $5$ms\\
 \cline{3-4}
 & & $[64\times192]$ & $4$ms \\
 \hline
 \multirow{2}{*}{\texttt{INT8}-LN (scaled)} & \multirow{2}{*}{$96.10\%$} & $[256\times44]$ & $5$ms\\
 \cline{3-4}
 & & $[64\times192]$ & $4$ms \\
\hline
\end{tabular}}
\end{center}
\end{table}

To identify the bottleneck in the inference process, we evaluate the runtime breakdown for different layers.
As depicted in Fig. \ref{fig:runtime-latency}(a), our findings indicate that the \texttt{Softmax} operator is responsible for the majority of the inference time.
Within the \texttt{Softmax} operator, the most time-consuming step is the negative exponential calculation.
We discovered that as the input size increases, the results of the negative exponential can be reused.
Leveraging this observation, SparseEngine employs a bitmap lookup table to reduce redundant calculations and optimize the \texttt{Softmax} operator.
Fig. \ref{fig:runtime-latency}(b) provides a comparison of the inference time for the main operators between CMSIS-NN and SparseEngine.
\textcolor{black}{Since the dense model can not be deployed on MCU using CMSIS-NN, to make a fair comparison, we use CMSIS-NN to perform sparse inference with the same algorithm as SparseEngine in Fig. \ref{fig:fig-tinyformer-sparseconv}, but without the SIMD acceleration.
The setting allows all the engines has the same memory usage when calculating the same operators.
The algorithm first decodes the sparse weight to get the weight information.
After decoding, the algorithm extracts the corresponding sub-matrix from the activation map, transforming it to column and perform the multiple-add calculation.}
Compared with CMSIS-NN, SparseEngine achieves $1.33\times$ to $1.55\times$ acceleration rate in \texttt{Conv2d} and \texttt{Linear} operator, and $22.5\times$ acceleration in \texttt{Softmax} operator.

\textcolor{black}{Moreover, we conducted a comparative analysis of power consumption between CMSIS-NN and SparseEngine on the STM32F746. 
Under the same voltage, the average power consumption of TinyFormer was measured at 473 mW, with a 3\% reduction compared to the average power consumption of CMSIS-NN. 
Adoption of a lookup table for softmax cut down the amount of calculation, and results in a slight reduction on power consumption.
Although the power reduction effect of SparseEngine is not particularly significant, SparseEngine achieve much less inference time compared to CMSIS-NN.
The total power consumption of SparseEngine for image recognition tasks was only 14\% of that of CMSIS-NN. 
The analyzing results suggest that SparseEngine holds a strong advantage in the TinyML applications that use transformers and require low-energy consumption.}


Additionally, we conduct an experiment involving Scaled-LayerNorm in TinyFormer.
Scaled-LayerNorm executes integer-arithmetic calculations, making it more suitable for model inference on MCUs.
Scaled-LayerNorm addresses the issue of precision loss in quantization by expanding the range of normalization results during the rounding operation.
Tab. \ref{tab:layernorm} illustrates the impact of Scaled-LayerNorm on acceleration.
Remarkably, the accuracy of TinyFormer using Scaled-LayerNorm is nearly equivalent to that of the normal LayerNorm computed in \texttt{FP32} format, while the inference procedure is accelerated by a factor of $38.8\times$ to $52.0\times$.
These experimental results align with our expectations, as Scaled-LayerNorm significantly enhances the efficiency of LayerNorm inference without any noticeable loss in accuracy.
In summary, the experimental results for both \texttt{Softmax} and Scaled-LayerNorm validate our observations and highlight the substantial acceleration effects achieved by SparseEngine.

%% file: 5-conclusion.tex
\section{Conclusions} \label{sec:conclusion}

In this work, TinyFormer is proposed as an innovative framework for developing efficient transformers on MCUs by integrating SuperNAS, SparseNAS, and SparseEngine.
One notable feature of TinyFormer is its ability to produce sparse models with high accuracy while adhering to hardware constraints.
By integrating model sparsity and neural architecture search, TinyFormer achieves a delicate balance between efficiency and performance.
Along with the automated deployment approaches, TinyFormer can further accomplish efficient sparse inference with a guaranteed latency on targeted MCUs.
Experimental results demonstrate that TinyFormer could achieve $96.1\%$ accuracy on CIFAR-10 under the limitations of $1$MB storage and $320$KB memory. 
Compared with CMSIS-NN, TinyFormer achieves a remarkable speedup of up to $12.2\times$ and reduces storage requirements by up to $78\%$.
These achievements not only bring powerful transformers into TinyML scenarios but also greatly expand the scope of deep learning applications.